# MULTIGRID TRAINING FOR MOLECULAR GENERATION USING GRAPH NEURAL NETWORKS

ZIXUAN LING *, PAULA MERCURIO †, AND DI LIU ‡

**Abstract.** Deep learning has demonstrated significant success for modeling bio-chemical molecular systems, where inputs are commonly represented as graphs or 3D grids. A major challenge is that computational cost scales with resolution, making full graph/grid computation of molecular densities expensive and often unstable. We introduce a multigrid training strategy that leverages low resolution optimization to accelerate learning at higher resolution through parameter transfer across discretizations. For graph molecular representations, we progressively transfer parameters learned from a coarse graph to a sequence of increasingly finer graphs via biased random walk upsampling. For 3D molecular generation, we voxelize the molecular structures at multiple resolutions, pretrain a coarse resolution conditional Variational Autoencoder (CVAE), and initialize a fine resolution CVAE by transferring shape compatible convolutional parameters from the coarse model. Numerical experiments on receptor-conditioned 3D Ligand generation show that multigrid training accelerates convergence and improves generalization compared to training from scratch.

## 1. Introduction.

The chemical space of stable, synthetically accessible molecules is enormous, but only a small fraction consists of promising drug candidates [42, 51, 8]. In practice, drug discovery proceeds through repeated rounds of proposed candidates and evaluations. Given the high cost of large scale experimental testing, the field is increasingly relying on computational methods to generate and prioritize candidates based on in-silicon predicted properties. The growing availability of public structural data, including protein–ligand complexes in the Protein Data Bank [2] and curated structure affinity resources such as PDBbind [28] and Binding MOAD [1], together with large scale bio-activity repositories such as BindingDB [27] and ChEMBL [12], have also made data driven molecular modeling more practical.

For modeling molecular structures, Deep neural networks can learn rich nonlinear mappings from input. For example, AlphaFold2 predicts the three-dimensional structure of a protein directly from its amino acid sequence [22]. Analogous predictive models have also been widely used to estimate molecular properties (e.g., drug-likeness or synthetic accessibility) from molecular representations [3, 54, 9]. Recent work has increasingly shifted from prediction to molecular design, where the goal is to generate novel molecules that satisfy preferred properties.

A widely adopted generative model in molecular design is the Variational Autoencoder (VAE), which learns a continuous latent representation that supports sampling and interpolation. The latent space provides a convenient interface for generating diverse candidates and for steering generation toward specific properties via conditional or guided sampling. These benefits come at a computational cost: training and sampling become substantially more expensive as the resolution of the molecular representation increases, i.e. larger graphs and finer voxel grids, which motivates multilevel strategies that iterate between coarse and fine models to efficiently solve the multiscale system.

*Department of Mathematics, Michigan State University, East Lansing, MI. (lingzixu@msu.edu)
†Department of Mathematics, Hamline University, Saint Paul, MN. (pmercurio01@hamline.edu)
‡Department of Mathematics, Michigan State University, East Lansing, MI. (liudi1@msu.edu)

Deep generative models for molecular design often formulate generation as a sequence modeling problem over SMILES (i.e. Simplified Molecular Input Line Entry System) strings. Each molecule is represented as a token sequence $x_i = (s_{i,1}, \ldots, s_{i,T_i})$ with $s_{i,t} \in \mathcal{V}$, where $\mathcal{V}$ is a token vocabulary of SMILES. Such sequences are modeled with a variational encoder-decoder architecture, typically with a recurrent decoder that generates tokens autoregressively [13]. Several approaches improve string validity by incorporating explicit syntactic constraints during learning and decoding [24, 7]. Sequence based generators have also been fine tuned with reinforcement learning to bias sampling toward property driven objectives [33].

Molecules can also be represented as graphs such that each sample has the form $x_i = (V_i, E_i)$, where each atom $v \in V_i$ carries discrete attributes such as element type, aromaticity, and formal charge, while each bond $e \in E_i$ is labeled by its bond type. Graph Neural Networks (GNNs) learn from such data by passing messages along edges and aggregating neighborhood information to obtain node embeddings and a graph level representation [17, 48]. Many graph based generators also adopt an encoder-decoder design. For example, GraphVAE encodes an input molecular graph into a latent representation and decodes by predicting an adjacency structure together with node and edge labels [44]. Chemical validity is enforced by decoding via constrained action spaces and using reinforcement learning to bias generation toward target properties [5, 57]. Other approaches generate molecules by composing chemically significant substructures into a full molecular graph [21]. Normalizing flow models have been developed for molecular graphs, enabling exact likelihood evaluation and efficient sampling [30]. Despite these advances, standard message passing architectures are largely flat and do not explicitly encode multiscale organization. To introduce hierarchy, several works, including DiffPool and Graph U-Nets [56, 11], augment GNNs with learnable pooling or coarsening modules. Later variants refine pooling by improving node selection and better preserving locality [25, 40].

While 2D graphs capture bonding relations, structure based applications such as drug discovery are inherently 3-dimensional. One example is receptor-conditioned 3D Ligand generation, where the goal is to identify small molecules (ligands) that adopt plausible binding poses and form favorable interactions within the binding pockets of target proteins (receptors). In the 3D setting, each ligand is represented by $x_i = \{(a_{ij}, r_{ij})\}_{j=1}^{N_i}$, where $N_i$ is the number of atoms in the ligand, $a_{ij}$ gives the atom type and $r_{ij} \in \mathbb{R}^3$ denotes the Cartesian coordinates. Before pocket-conditioned 3D generation became practical, early deep learning work in structure-based drug design focused on scoring: learning functions that recognize favorable protein-ligand interaction patterns. A common strategy voxelizes the receptor environment and candidate poses in multi-channel 3D atomic grids and scores them with 3D convolutional neural networks [50, 37]. Such learned scorers have been integrated into docking by combining conventional pose search with neural scoring [32]. The grid formulation also provides differentiable objectives that support gradient based pose refinement [39, 46] and analysis that highlight which features of the pocket-ligand most influence the predicted score [18].

Building on voxel grid scoring models, early pocket-conditioned generators also operated on 3D grids, either by generating ligand density grids aligned to the binding site and then performing atom fitting and bond inference [38], or by learning a conditional score on voxel grids and sampling with Langevin dynamics [36, 49]. More recent methods move beyond grids to direct 3D graph/point-cloud formulations and enforce equivariance to translations and rotations [19]. A prominent approach is the gener-ation of autoregressive pocket-conditioned ligands, which grows a ligand sequentially by placing atoms in a 3D space conditioned on the pocket and the current partial ligand [29, 35, 26]. Subsequent work strengthens stepwise conditioning by explicitly encoding pocket-ligand interaction roles, improving control over contact patterns, and generalizing to unseen targets [58]. Diffusion models have also been widely used for 3D ligand generation [43, 15, 16, 34, 31], although their iterative sampling can make inference more expensive than one-shot decoders such as VAEs.

Recent work has explored hierarchical latent variable VAEs on atomic density grids, using hierarchical priors to better capture multiscale molecular structures [52]. However, these latent levels do not explicitly correspond to a prescribed hierarchy of grid resolutions [45, 47, 6, 41]. From a computational perspective, both graph message passing and voxel grid generative modeling become increasingly expensive as resolution grows. This suggests connections to multigrid methods that have been very effective for broad computational tasks, from numerical solutions of Partial Differential Equations [4] to large scale nonlinear optimization [53]. Multigrid approaches construct a hierarchy of discretizations from coarse to fine, couple different levels through transfer operators such as restriction and prolongation, and use coarse grid corrections to reduce low frequency error components on finer levels. Coarse levels provide inexpensive approximations that supply effective trial points, thereby accelerating convergence on the fine problem.

In this paper, we adapt the classical multigrid methodology to resolution dependent deep learning for molecular modeling. Specifically, we make the following contributions:

- We propose a multigrid training framework for GNNs that builds a hierarchy of graphs through biased random walk upsampling and transfers learned parameters across levels. We analyze cross-level propagation consistency and show that under overlap alignment and sufficient coverage, the transferred weights provide a provable well conditioned start with bounded fine level loss and gradient norm at initialization.
- For pocket conditioned 3D ligand generation, we construct receptor-ligand density grids at multiple voxel resolutions within a fixed binding site box and initiate the fine resolution conditional variational autoencoder (CVAE) by transferring convolutional parameters from the coarse model.
- We empirically compare multigrid training against single resolution training from scratch, quantifying improvements in training efficiency and generalization for graph learning, as well as ligand quality for receptor-conditioned generation.

## 2. Background.

In this section, we briefly review the graph learning setting used in our multigrid training framework, focusing on message passing graph neural networks (GNNs) and standard supervised objectives for node and graph classification.

### 2.1. Graph Neural Network.

Let $G = (V, E)$ denote an undirected graph of $|V| = N$ nodes with adjacency matrix $A = \{a_{ij}\} \in \{0,1\}^{N\times N}$ and node feature matrix $X \in \mathbb{R}^{N\times D}$, where each node is associated with a vector of dimensions $D$. The goal of node embedding is to learn a function $f : V \to \mathbb{R}^d$ that assigns a label $y_i$ to each node $v_i$ based on the overall structure of the graph and the note features. The learned embedding function $y = f(v)$ is optimized to reflect the proximity of nodes in the original graphs and networks, and therefore it can be used for downstream machine learning tasks such as clustering and classification.

We build on the framework of message passing graph neural networks, which update node embeddings through iterative aggregation of neighborhood information. A generic $L$-layer message passing GNN updates node embeddings according to the following recursion:

$$H^{(l)} = M(A, H^{(l-1)}; \theta^{(l)}), \quad l = 1, 2, \ldots, L, \tag{2.1}$$

where $H^{(l)} \in \mathbb{R}^{N \times D}$ denotes the node embeddings at $l$th layer, initialized with $H^{(0)} = X$. The function $M$ is a differentiable message propagation function that depends on the adjacency matrix $A$, the embeddings of the previous layer, and the learnable parameters $\theta^{(l)}$. Various GNN architectures differ in how $M$ is instantiated. A notable example is the Graph Convolutional Network (GCN) [23], which uses the propagation rule:

$$H^{(l)} = \sigma\left(\tilde{A} H^{(l-1)} W^{(l-1)}\right), \tag{2.2}$$

where $\tilde{A} = D^{-\frac{1}{2}}(A + I)D^{-\frac{1}{2}}$ is the symmetrically normalized adjacency matrix with added self-loops, $W^{(l-1)} \in \mathbb{R}^{D \times D}$ is a trainable weight matrix and $\sigma(\cdot)$ is a non-linearity such as ReLU. Our multigrid framework is applicable to general message passing GNNs of the form given in Equation (2.1) and does not rely on the specific implementation of $M$.

**2.2. Supervised Classification.** We apply a softmax classifier row-wise to the final node embeddings $H^{(L)} \in \mathbb{R}^{N \times F}$, where $F$ is the number of classes. The predicted class distribution for node $v_i$ is given by:

$$Z_i = \mathrm{softmax}(H_i^{(L)}), \quad \text{where} \quad \mathrm{softmax}(z_j) = \frac{\exp(z_j)}{\sum_{f=1}^{F} \exp(z_f)}.$$

Given a set of labeled nodes $\mathcal{V}_L \subset V$, the model is trained to minimize the cross-entropy loss over this labeled subset:

$$\mathcal{L} = -\sum_{l \in \mathcal{V}_L} \sum_{f=1}^{F} Y_{lf} \log Z_{lf},$$

where $Y_{lf} \in \{0, 1\}$ is the one-hot encoded ground-truth label for node $v_l$, and $Z_{lf}$ is the predicted probability of class $f$. The model parameters $\{\theta^{(k)}\}_{k=1}^{K}$ (e.g. $W^{(k)}$ in the GCN case) are optimized using gradient descent.

**2.3. Graph Representation.** In addition to node level tasks, many applications such as molecular generation require learning representations for entire graphs. In graph classification, the input consists of a set of graphs $\mathcal{G} = \{G_1, G_2, \ldots, G_n\}$, where each graph $G_i = (V_i, E_i)$ has its own adjacency matrix $A_i$, node features $X_i \in \mathbb{R}^{|V_i| \times D}$. The goal is to learn a function $f : \mathcal{G} \to \mathcal{Y}$ that maps each graph to a class label $y \in \mathcal{Y}$. To do so, message passing GNNs are first applied to compute node embeddings as in Equation (2.1). A permutation invariant pooling function (e.g. global mean, sum, or max pooling) is then used to aggregate the node embeddings into a graph embedding. This embedding is passed to a classifier to predict the graph label. As with node classification, the model is trained to minimize the cross-entropy loss over a labeled training set of graphs.

**3. Multigrid GNN.** In this section, we introduce the multigrid training framework for graph neural networks. We first define a graph upsampling operator that builds a hierarchy of increasingly refined subgraphs and then describe how we transfer and update model parameters across levels in a multigrid training procedure.

**3.1. Graph Upsampling.** We construct the multigrid framework as a hierarchy of graphs, where each level corresponds to a different resolution. This requires an upsampling mechanism that systematically expands a coarse graph into a finer one. In structured grid based methods, upsampling is typically performed using interpolation. However, these operations do not directly apply to graph data due to their irregular connectivity. We propose an upsampling method based on biased random walks, inspired by [14]. Starting from the coarse graph, we explore connectivity and add newly visited nodes and edges to form the next level.

Formally, we propose the layerwise upsampling rule for a coarse subgraph $G_k = (V_k, E_k, X_k)$ as

$$G_{k+1} = \mathcal{U}\big(G_k; \ell, w, \alpha\big),$$

where $\ell$ is the length of the random-walk, $w$ is the number of walks per node, and $\alpha$ is a bias weight toward unexplored neighbors. The upsampling operator $\mathcal{U}$ expands $G_k$ to $G_{k+1} = (V_{k+1}, E_{k+1}, X_{k+1})$ by exploring the original full graph during random walks. For a chosen start node $u \in V_k$, we simulate a walk $\{c_0, c_1, \dots, c_{\ell-1}\}$ with $c_0 = u$. At each step $i$, the next node $c_{i+1}$ is chosen from the neighbors of the current node $c_i$ according to a biased transition distribution:

$$P\big(c_{i+1} = x \mid c_i = v\big) = \begin{cases} \frac{\pi_{vx}}{Z_v} & \text{if } (v,x) \in E_k, \\ 0 & \text{otherwise}, \end{cases}$$

where $\pi_{vx}$ is an unnormalized weight that encodes our bias toward visiting previously unvisited nodes, and $Z_v = \sum_{(v,x)\in E_k} \pi_{vx}$ is the normalizing constant. We set

$$\pi_{vx} = \begin{cases} \alpha, & \text{if } c_j \neq x \text{ for all } 0 \leq j \leq i, \\ 1.0, & \text{otherwise.} \end{cases}$$

After performing $w$ biased walks from each node in $V_k$, we group all newly visited nodes into $V_{k+1}$ and add any newly encountered edge $(v, x)$ into $E_{k+1}$. The feature matrix $X_{k+1}$ should inherit the row vectors of the original feature set $X$.

The computational cost of the upsampling procedure depends primarily on the number of nodes $|V_k|$ in the current coarse graph $G_k$, the length of the random walk $\ell$, the number of walks per node $w$, and the average degree of the node $d$ in $G_k$. Each walk step involves sampling a neighbor based on a biased transition probability, leading to an overall runtime of $\mathcal{O}(w\,\ell\,d\,|V_k|)$. This runtime scales linearly with $|V_k|$, and since $d$ is typically small for real world networks, the procedure remains efficient for hierarchical training. Moreover, since the random walks originating from different nodes are independent, the procedure can be executed efficiently in parallel on GPUs.

**3.2. Multigrid Training.** The gist of multigrid algorithms is to leverage information from the coarser levels to generate new trial points for problems on finer grids, thereby improving efficiency while preserving essential structural details. Inspired by this principle, we propose a multigrid GNN framework that progressively refines the graph from a coarse level $G_0$ to the original input graph $G$ for node classification tasks.

*Coarse Graph Sampling and Training.* We begin by sampling a small subgraph $G_0$ from the original graph $G$. This is done by selecting multiple random starting nodes to form an initial subgraph $G_{\text{tiny}}$ without edges. We apply the upsampling operator $\mathcal{U}$ to obtain $G_0 = \mathcal{U}(G_{\text{tiny}})$, producing a reduced graph that captures key connectivity patterns while allowing efficient training. Let $X_0 \in \mathbb{R}^{|V_0| \times D}$ denote the node features restricted to $V_0$. We train a GNN on $G_0$ to obtain initial weights $\{W_0^l\}_{l=1}^L$, where each $W_0^l \in \mathbb{R}^{D^l \times D^l}$ is the learned weight matrix at layer $l$. These parameters form the initial conditions for subsequent refinements. We then upsample $G_0$ to form a finer graph $G_1$, incorporating additional structural details for next level training.

*Parameter Sharing.* We first adopt a parameter sharing approach by reusing GNN weights trained on coarser graphs to initialize models on finer graphs. After training a GNN on the coarse graph $G_0$, we obtain weight matrices $\{W_0^l\}_{l=1}^L$, which are then used to initialize the corresponding layers on the finer graph $G_1$; that is, we set $W_1^l = W_0^l$ for each layer $l = 1, \ldots, L$. This process is repeated as we upsample to successive levels $G_2, G_3, \ldots$, with each new level initialized using the trained parameters from the previous one. Once we reach the final level $G_K = G$, no further upsampling is necessary and the model is trained on all nodes and edges of the original graph. Although the graphs differ in connectivity, the embedding dimensions $D^l$ remain fixed, allowing the parameters to be transferred directly without architectural changes. This multigrid refinement allows each finer level GNN to benefit from a well informed initialization, accelerating convergence, and preserving knowledge learned at coarser resolutions.

*Layer Freezing.* As an alternative to parameter sharing, we also adopt a coarse-to-fine training strategy that progressively freezes layers to retain the knowledge learned at earlier stages. After training a GNN on the coarse graph $G_0$, we obtain weight matrices $\{W_0^l\}_{l=1}^L$, which are transferred to initialize the corresponding layers on the finer graph $G_1$. During training on $G_1$, we freeze the lower layers (e.g., the first $m_0$ layers) and update only the remaining layers to adapt to the finer structure. This process is repeated as we upsample to successive levels $G_2, G_3, \ldots$, where at each level more layers are frozen to preserve previously learned representations. This strategy allows earlier layers to capture coarse level structural patterns, while deeper layers are progressively fine tuned to the detailed topology of finer graphs. In doing so, the model gradually shifts its focus from global to local information.

*Self-Avoiding Mechanism.* Both parameter sharing and layer freezing strategies can be improved by a self-avoiding mechanism that restricts each intermediate training stage to newly added nodes only. After training on the coarse graph, we proceed to train on a finer subgraph, excluding the nodes already seen during the coarse stage. This self-avoiding scheme ensures that each stage trains on a disjoint subset of nodes, thereby enforcing a hierarchical decomposition of the learning task. Conceptually, this scheme resembles the bottom-up pass of a full multigrid cycle. The final training phase on the full original graph, with all layers unfrozen, acts as a top level correction step that integrates the multi-resolution representations. The complete multigrid training procedure is summarized in Algorithm 3.1.

**3.3. Cross-level Consistency.** Although our multigrid GNN framework trains models sequentially from coarse to fine levels, it is important to maintain cross-level consistency, which is a well-established strategy for improving convergence and efficiency in classical multigrid methods.

To this end, we introduce a guidance loss to promote consistency across levels. After training on a coarse subgraph, we store its predictions. When training on the

**Algorithm 3.1** General Multigrid GNN Framework

**Input:** Graph $G = (V, E)$, feature matrix $X$, GNN model, upsampling operator $\mathcal{U}$, number of levels $K$, number of layers $L$, optional freezing schedule $\{m_k\}_{k=0}^{K-1}$.
Initialize $G_{\text{tiny}}$ by sampling a set of nodes from $V$ (with no edges).
Form the initial coarse graph $G_0 \leftarrow \mathcal{U}(G_{\text{tiny}})$.
Train the GNN on $G_0$ to obtain weights $\{W_0^l\}_{l=1}^L$.
**for** $k = 0, 1, \ldots, K-1$ **do**
  Upsample: $G_{k+1} \leftarrow \mathcal{U}(G_k)$.
  Initialize parameters on level $k{+}1$ by transfer: $W_{k+1}^l \leftarrow W_k^l$ for $l = 1, \ldots, L$.
  **if** layer freezing is used **then**
    Freeze layers $1, \ldots, m_k$ and fine tune layers $m_k{+}1, \ldots, L$ on $G_{k+1}$.
  **if** self-avoiding mechanism is used **then**
    Define the training node set $\mathcal{V}_{k+1}^{\text{train}} \leftarrow V_{k+1} \setminus V_k$.
  **else**
    Define the training node set $\mathcal{V}_{k+1}^{\text{train}} \leftarrow V_{k+1}$.
  Train the GNN on $G_{k+1}$ using labeled nodes in $\mathcal{V}_{k+1}^{\text{train}}$.
Unfreeze all layers and train the GNN on the full graph $G_K = G$ using labeled nodes in $V$.
**Output:** Final weights $\{W_K^l\}_{l=1}^L$.

next finer graph, we add an extra loss term that penalizes discrepancies between the current predictions and the stored outputs on overlapping nodes. This encourages the finer-level model to align with the coarse-level solution while adapting to the finer graph structure such that

$$\mathcal{L} = \mathcal{L}_{\text{GNN}} + \lambda \mathcal{L}_{\text{Guidance}}, \tag{3.1}$$

where $\mathcal{L}_{\text{GNN}}$ is the standard supervised loss at the current level, $\mathcal{L}_{\text{Guidance}}$ is the cross-level consistency term, and $\lambda \geq 0$ controls its strength. We define the guidance loss at level $k{+}1$ as the cross-entropy between the stored predictions from level $k$ and the current predictions at level $k{+}1$:

$$\mathcal{L}_{\text{Guidance}} = -\sum_{l \in V_k} \sum_{f=1}^{F} \hat{y}_{lf}^{(k)} \log y_{lf}^{(k+1)}, \tag{3.2}$$

where $\hat{y}_{lf}^{(k)}$ is the stored probability that node $v_l$ belongs to class $f$ at level $k$, and $y_{lf}^{(k+1)}$ is the predicted probability that node $v_l$ belongs to class $f$ at level $k+1$.

**4. Analysis of the Multigrid Scheme.** We analyze our multigrid training scheme by quantifying cross-level propagation consistency on the overlap set. In particular, we derive bounds on the prediction, loss, and gradient gaps between the coarse and fine objectives, which imply that coarse-to-fine transfer provides a warm start under good alignment and high coverage. Throughout this section, we assume the activation function is Lipschitz and has a Lipschitz derivative.

ASSUMPTION 4.1. *The activation $\sigma : \mathbb{R} \to \mathbb{R}$ is $L_\sigma$-Lipschitz, i.e., $|\sigma(a) - \sigma(b)| \leq L_\sigma |a - b|$ for all $a, b \in \mathbb{R}$. Moreover, its derivative $\sigma'$ is $L_{\sigma'}$-Lipschitz, i.e., $|\sigma'(a) - \sigma'(b)| \leq L_{\sigma'} |a - b|$ for all $a, b \in \mathbb{R}$.*

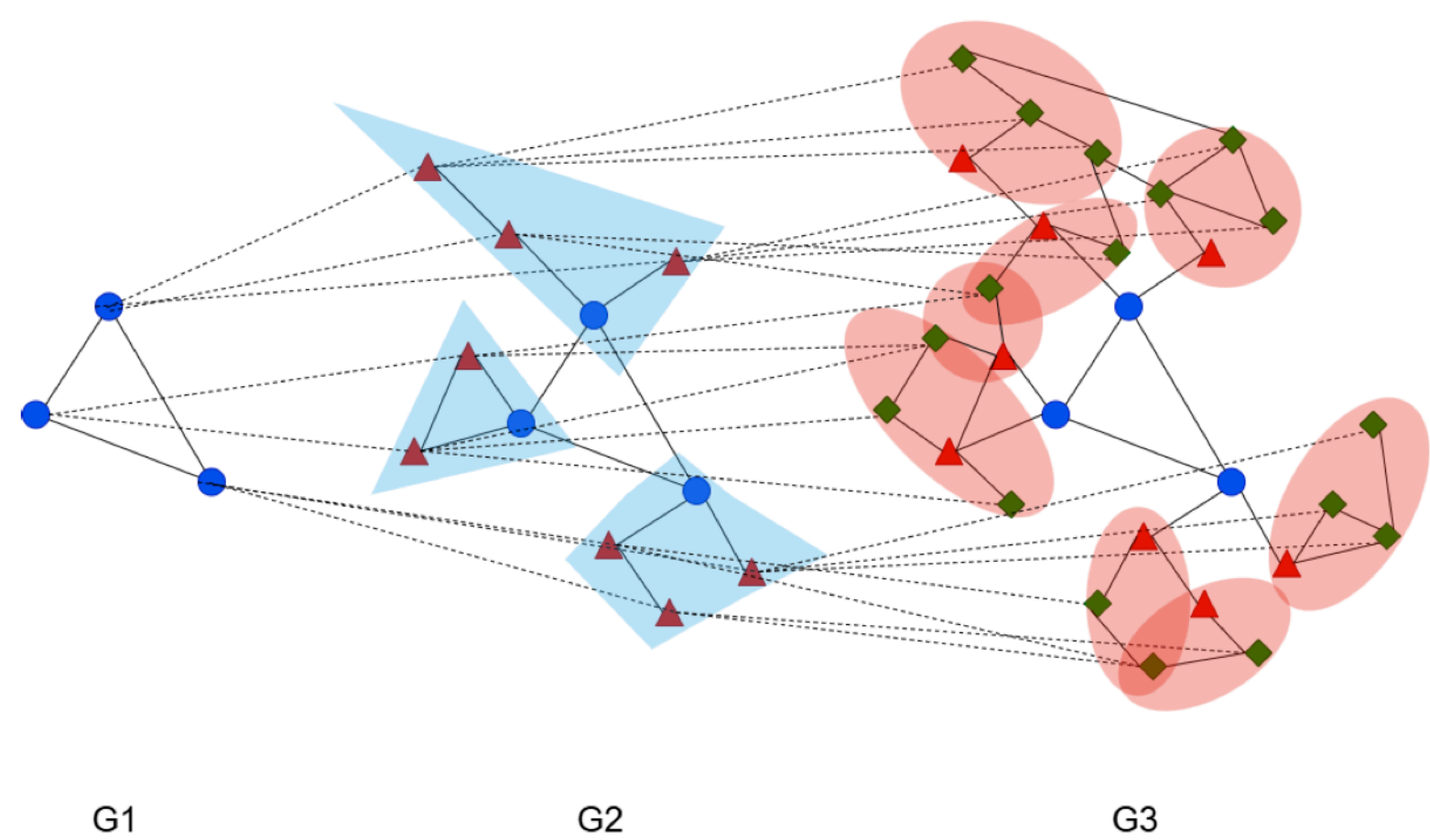

Fig. 1: Three-level coarse-to-fine hierarchy. In $G_1$, blue circles denote the coarse-level nodes. In $G_2$, new nodes appear as red triangles, and blue regions indicate areas upsampled from the corresponding blue circle in $G_1$. In $G_3$, green squares indicate the fine-level nodes, and red ellipses indicate the neighborhoods upsampled from the corresponding red triangles in $G_2$.

We apply $\sigma$ and $\sigma'$ entrywise: for $z \in \mathbb{R}^n$, $(\sigma(z))_i = \sigma(z_i)$ and $(\sigma'(z))_i = \sigma'(z_i)$.

*Remark* 4.2. Under Assumption 4.1, for any $z, z' \in \mathbb{R}^n$,

$$\|\sigma(z) - \sigma(z')\|_2 \leq L_\sigma \|z - z'\|_2.$$

Indeed, by Assumption 4.1, for each $i$ we have $|\sigma(z_i) - \sigma(z_i')| \leq L_\sigma |z_i - z_i'|$. Squaring and summing over $i$ yields

$$\|\sigma(z) - \sigma(z')\|_2^2 = \sum_{i=1}^{n} |\sigma(z_i) - \sigma(z_i')|^2 \leq L_\sigma^2 \sum_{i=1}^{n} |z_i - z_i'|^2 = L_\sigma^2 \|z - z'\|_2^2,$$

and taking square roots gives the claim. The proof for $\sigma'$ is identical, giving $\|\sigma'(z) - \sigma'(z')\|_2 \leq L_{\sigma'} \|z - z'\|_2$.

**4.1. Coarse–Fine Training Setup.** We analyze a single layer GCN for node-wise regression with squared loss, a minimal setup capturing the essential transfer effect. The classification case will be similar. For simplicity, we present the scalar output case $p = 1$ (so $w \in \mathbb{R}^d$). The multi-output case with $W \in \mathbb{R}^{d \times p}$ follows by applying $\sigma$ entry-wise and measuring residuals and gradients with the Frobenius norm.

We work with a fine graph $G = (V, E)$ of $n$ nodes and a feature matrix $X \in \mathbb{R}^{n \times d}$. The coarse graph $G_c = (U, E_c)$ is a subgraph with $U \subset V$, $|U| = m$, and $E_c \subseteq E$. To relate the two node sets, let $Q \in \{0, 1\}^{n \times m}$ be the selection matrix for $U$, obtained by taking $m$ columns of the identity $I_n$ indexed by $U$. Then $X_c := Q^\top X \in \mathbb{R}^{m \times d}$ collects coarse features and $Y_c := Q^\top Y \in \mathbb{R}^m$ extracts coarse targets from $Y \in \mathbb{R}^n$. We write $A \in \mathbb{R}^{n\times n}$ and $A_c \in \mathbb{R}^{m\times m}$ for the self-loop augmented, symmetrically normalized adjacency matrices of $G$ and $G_c$, respectively.

We use a single layer GCN with weight vector $w \in \mathbb{R}^d$ and a pointwise nonlinearity $\sigma : \mathbb{R} \to \mathbb{R}$. On the fine graph we compute predictions $\hat{Y}_f(w) = \sigma(AXw)$, and on the coarse graph we have $\hat{Y}_c(w) = \sigma(A_c X_c w)$. We measure fit with a node-averaged squared loss. With residuals $r_f = \hat{Y}_f - Y$ and $r_c = \hat{Y}_c - Y_c$, the loss functions for the fine and coarse graphs are $L_f(w) = \frac{1}{2n}\|r_f\|_2^2$ and $L_c(w) = \frac{1}{2m}\|r_c\|_2^2$, respectively.

On the overlap $U$, we compare two routes for moving information across levels: propagate on the fine graph $G$ and then restrict to $U$, or restrict to $U$ first and then propagate on the coarse graph $G_c$. Their discrepancy is measured by the operator $E := Q^\top A - A_c Q^\top \in \mathbb{R}^{m\times n}$ and its spectral norm $\delta := \|E\|_2$. A small $\delta$ means the coarse operator is well aligned with the fine operator on the shared nodes. For $\delta$ to be meaningful on graphs, it should not depend on how we label the nodes. The following proposition shows that relabeling $V$ or $U$ leaves $\delta$ unchanged.

PROPOSITION 4.3 (Permutation invariance). *Suppose $P \in \{0,1\}^{n\times n}$ and $R \in \{0,1\}^{m\times m}$ are permutation matrices acting on $V$ and $U$, respectively. Define*

$$A' = PAP^\top, \qquad A'_c = RA_cR^\top, \qquad Q' = PQR^\top.$$

*Then*

$$\left\| Q'^\top A' - A'_c Q'^\top \right\|_2 = \left\| Q^\top A - A_c Q^\top \right\|_2 = \delta.$$

*Proof.* We compute

$$Q'^\top A' - A'_c Q'^\top = (RQ^\top P^\top)(PAP^\top) - (RA_cR^\top)(RQ^\top P^\top) = R\,(Q^\top A - A_c Q^\top)\,P^\top.$$

Since $P$ and $R$ are orthogonal, $\|RMP^\top\|_2 = \|M\|_2$ for any $M$. Therefore

$$\left\|Q'^\top A' - A'_c Q'^\top\right\|_2 = \left\|Q^\top A - A_c Q^\top\right\|_2 = \delta. \qquad \square$$

*Remark* 4.4. To see why we set $Q' = PQR^\top$, let $x \in \mathbb{R}^n$ be a signal on the fine nodes $V$. Restricting $x$ to the coarse set $U$ gives $Q^\top x \in \mathbb{R}^m$. After relabeling, the fine signal becomes $Px$ and the coarse signal becomes $R(Q^\top x)$. We require restriction to commute with relabeling:

$$(Q')^\top (Px) = R\big(Q^\top x\big) \qquad \text{for all } x \in \mathbb{R}^n.$$

Hence $(Q')^\top P = RQ^\top$, which implies $Q' = PQR^\top$.

Hence, we may treat $\delta$ as an intrinsic graph quantity for the pair $(G, G_c)$ over $U$, invariant to relabeling.

**4.2. Prediction Mismatch and and Loss Gap.** We begin by quantifying the discrepancy between the fine-level and coarse-level predictions on the shared node set $U$. This prediction mismatch is the basic quantity that drives the subsequent comparison of the two objectives.

LEMMA 4.5 (Prediction mismatch on $U$). *For any $w \in \mathbb{R}^d$, we have*

$$\left\| Q^\top \hat{Y}_f(w) - \hat{Y}_c(w) \right\|_2 \le L_\sigma\, \delta\, \|X\|_2\, \|w\|_2.$$

*Proof.* Because $Q^\top$ selects coordinates and $\sigma$ acts entrywise, the two operations commute: $Q^\top\sigma(z) = \sigma(Q^\top z)$. Using $Q^\top AXw - A_cX_cw = EXw$, we get

$$\begin{aligned}\|Q^\top \hat{Y}_f(w) - \hat{Y}_c(w)\|_2 &= \|\sigma(Q^\top AXw) - \sigma(A_cX_cw)\|_2 \\ &\le L_\sigma \|EXw\|_2 \\ &\le L_\sigma\,\delta\,\|X\|_2\,\|w\|_2.\end{aligned}$$

□

This bound shows that the prediction discrepancy on $U$ is controlled by the alignment parameter $\delta$ and scales linearly with the activation Lipschitz constant $L_\sigma$, the feature norm, and the weight norm. The dependence on $\|X\|_2\,\|w\|_2$ suggests that standard training techniques such as feature normalization and weight decay tighten the bound.

We now quantify how the training objectives on the two levels relate.

THEOREM 4.6 (Loss gap). *Let $\rho := m/n$ be the coarse-to-fine node ratio. Define $r_f^U := Q^\top r_f$ and $r_f^{U^c} := (I - QQ^\top)r_f$, where $r_f^U$ is the fine residual restricted to $U$ and $r_f^{U^c}$ is the fine residual on $V \setminus U$. Then for any $w \in \mathbb{R}^d$,*

$$\big|L_f(w) - L_c(w)\big| \;\le\; \frac{\rho\, L_\sigma\, \delta}{2m}\,\big(\|r_f^U\|_2 + \|r_c\|_2\big)\,\|X\|_2\,\|w\|_2 \;+\; \frac{1-\rho}{2m}\,\|r_c\|_2^2 \;+\; \frac{1}{2n}\,\|r_f^{U^c}\|_2^2.$$

*Proof.* See Appendix A. □

The bound decomposes into an overlap term and two non-overlap corrections. The overlap term is controlled by the alignment parameter $\delta$ and scales with the feature and weight norms. The remaining two terms arise from averaging differences and residuals on nodes outside $U$ and disappear when $m = n$. Denote by $\varepsilon_{\text{loss}}(w)$ the right-hand side of the bound in Theorem 4.6. Then, for a well-trained coarse model, namely a point $w_c^\star$ with small $L_c(w_c^\star)$ and small $\|r_c(w_c^\star)\|_2$, Theorem 4.6 gives

$$L_f(w_c^\star) \;\le\; L_c(w_c^\star) + \varepsilon_{\text{loss}}(w_c^\star).$$

When $\delta$ is small and the coverage $\rho$ is close to 1, this discrepancy is small and decreases with the coarse residual $\|r_c(w_c^\star)\|_2$, so the transferred point achieves a comparable fine-level objective.

**4.3. Gradient Gap.** Next we compare the gradients of the fine and coarse objectives. To control operator norms in the gradient expressions, we first record a spectral bound for the symmetrically normalized adjacency matrix.

LEMMA 4.7 (Spectral norm of the normalized adjacency). *Let $\bar{A} \in \mathbb{R}^{n\times n}$ be the self-loop augmented adjacency matrix of an undirected graph, and let $D := \operatorname{diag}(\bar{A}\mathbf{1})$ be the corresponding degree matrix. Define the symmetrically normalized adjacency matrix associated with $\bar{A}$ by*

$$\tilde{A} \;:=\; D^{-1/2}\,\bar{A}\,D^{-1/2}.$$

*Then all eigenvalues of $\tilde{A}$ lie in $[-1, 1]$, and in particular $\|\tilde{A}\|_2 \le 1$.*

*Proof.* See Appendix B. □

Lemma 4.7 implies $\|A\|_2 \le 1$ and $\|A_c\|_2 \le 1$ for the propagation matrices used at the two levels, which we use in the proof of Theorem 4.8.

THEOREM 4.8 (Gradient gap). *Under the notation of Theorem 4.6, we have for any $w \in \mathbb{R}^d$,*

$$\left\|\nabla L_f(w) - \nabla L_c(w)\right\|_2 \le \frac{\delta}{n} \|X\|_2 \Big( L_\sigma \|r_f^U\|_2 + \big(L_\sigma^2 + L_{\sigma'} \|r_c\|_2\big) \|X\|_2 \|w\|_2 \Big) + \frac{L_\sigma}{n} \|X\|_2 \|r_f^{U^c}\|_2 + \left(\tfrac{1}{m} - \tfrac{1}{n}\right) L_\sigma \|X\|_2 \|r_c\|_2.$$

*Proof.* See Appendix C. ∎

Let $\varepsilon_{\text{grad}}(w)$ denote the right-hand side of Theorem 4.8. Then, for a well- trained coarse model $w_c^\star$ with $\|\nabla L_c(w_c^\star)\|_2 \approx 0$ and small $\|r_c(w_c^\star)\|_2$,

$$\|\nabla L_f(w_c^\star)\|_2 \le \varepsilon_{\text{grad}}(w_c^\star).$$

When $\delta$ is small and the coverage $\rho$ is close to 1, the term $\varepsilon_{\text{grad}}(w_c^\star)$ is small and decreases with the coarse residual. Thus the transferred point $w_c^\star$ is approximately first-order stationary for the fine objective.

**4.4. Multilevel Hierarchy.** We now extend the two-level mismatch operator $E = Q^\top A - A_c Q^\top$ to a $k$-level hierarchy. Let $G_\ell = (V_\ell, E_\ell)$, $\ell = 1, \ldots, k$, be a nested sequence of graphs with $V_1 \subset V_2 \subset \cdots \subset V_k$ and $|V_\ell| = n_\ell$. Let $A_\ell \in \mathbb{R}^{n_\ell \times n_\ell}$ be the (self-loop augmented) symmetrically normalized adjacency for $G_\ell$.

For each consecutive pair $(G_\ell, G_{\ell+1})$, let $Q_{\ell:\ell+1} \in \{0,1\}^{n_{\ell+1} \times n_\ell}$ be the selection matrix that embeds signals on $V_\ell$ into $V_{\ell+1}$ (equivalently, $Q_{\ell:\ell+1}^\top$ restricts a signal on $V_{\ell+1}$ to the shared nodes $V_\ell$). For $1 \le i < j \le k$, chaining selections across levels gives the composed map from $V_i$ to $V_j$,

$$Q_{i:j} = Q_{j-1:j}\, Q_{j-2:j-1} \cdots Q_{i:i+1} \in \mathbb{R}^{n_j \times n_i}.$$

Since $Q_{i:j}$ has orthonormal columns, $\|Q_{i:j}\|_2 = \|Q_{i:j}^\top\|_2 = 1$. We then define the cross-level propagation mismatch

$$E_{i,j} := Q_{i:j}^\top A_j - A_i Q_{i:j}^\top \in \mathbb{R}^{n_i \times n_j}, \qquad \delta_{i,j} := \|E_{i,j}\|_2.$$

As in Proposition 4.3, the mismatch operator $E_{i,j}$ and its norm $\delta_{i,j}$ are invariant to the relabeling of the nodes.

THEOREM 4.9 (Telescoping bound for multilevel mismatch). *For the $k$-level hierarchy above, we have*

$$\delta_{1,k} \le \sum_{\ell=1}^{k-1} \delta_{\ell,\ell+1}.$$

*Proof.* See Appendix D. ∎

Theorem 4.9 is a local-to-global consistency bound. It shows that the mismatch between the endpoints of the hierarchy is controlled by the accumulated mismatch incurred across adjacent level transitions. In particular, any prediction, loss, or gradient comparison that depends on the quantity of two-levels $\delta$ can be applied between $G_1$ and $G_k$ by replacing $\delta$ with the multilevel bound $\sum_{\ell=1}^{k-1} \delta_{\ell,\ell+1}$.

Now we specialize Theorem 4.9 under a geometric decay model for the adjacent-level mismatches.

THEOREM 4.10 (Bound under geometric mismatch decay). *Let $k \geq 2$. Suppose that there exist constants $C > 0$, $\alpha > 1$, and $q \in (0,1)$, and nonnegative scalars $\{\tau_\ell\}_{\ell=1}^{k-1} \subset [0,\infty)$ with fixed total $T := \sum_{\ell=1}^{k-1} \tau_\ell$, such that for each $\ell = 1,\dots,k-1$,*

$$\delta_{\ell,\ell+1} \leq C\tau_\ell^\alpha, \qquad \tau_\ell = \tau_1 q^{\ell-1}.$$

*Then*

$$\delta_{1,k} \leq C\tau_1^\alpha \frac{1-q^{\alpha(k-1)}}{1-q^\alpha} = CT^\alpha\Big(\frac{1-q}{1-q^{k-1}}\Big)^\alpha \frac{1-q^{\alpha(k-1)}}{1-q^\alpha}.$$

*Moreover, for fixed $T, C, \alpha, q$, the right-hand side decreases in $k$.*

*Proof.* See Appendix E. ∎

Theorem 4.10 shows that a chain of small, well-aligned transfers can be preferable to a single coarse-to-fine jump. In the geometric mismatch decay model, the refinement of the hierarchy (larger $k$) decreases the mismatch of the endpoint $\delta_{1,k}$. Consequently, the bounds in Lemma 4.5 and Theorems 4.6–4.8 tighten as intermediate levels are inserted.

**5. Generating Molecular Structures.** We further apply the multigrid strategy to study pocket-conditioned 3D ligand generation, that can find new ligand structures to form successful bonds with protein receptor sites. The method is based on recent development of generative molecular models, in the form of Variational Autoencoders (VAE), which learns a distribution of molecular structures parameterized by deep neural networks from a set of chemical datasets.

**5.1. Variational Autoencoders.** We first briefly recall the Variational Autoencoder (VAE) framework. Let $x \in \mathbb{R}^m$ denote an observed data sample and let $z \in \mathbb{R}^k$ be a latent variable with prior $p(z)$ (typically $p(z) = \mathcal{N}(0, I)$). A VAE consists of a decoder $p_\theta(x \mid z)$ and an encoder $q_\phi(z \mid x)$. Training maximizes the evidence lower bound (ELBO)

$$L_{\text{ELBO}}(\theta,\phi;x) := \mathbb{E}_{q_\phi(z|x)}\big[\log p_\theta(x \mid z)\big] - \text{KL}\big(q_\phi(z \mid x) \,\|\, p(z)\big),$$

where the first term encourages accurate reconstruction and the KL term regularizes the encoder toward the prior.

If side information $c$ is available (e.g., a receptor, in the 3D ligand generation setting), we use a conditional VAE with decoder $p_\theta(x \mid z, c)$ and encoder $q_\phi(z \mid x, c)$ and optimize the corresponding conditional ELBO. All multigrid constructions below apply to both VAEs and CVAEs. For brevity, we present the unconditional form.

**5.2. Multigrid Implementation.** We describe a coarse-to-fine, multiresolution training scheme for VAEs, where a model trained at a coarse resolution is used to initialize training at a finer resolution. Since the molecular structures are described by 3D coordinates, instead of network relations, we do not employ the random walk upsampling technique for generating multigrid hierarchy.

Given a dataset $X = \{x_i\}_{i=1}^N$, we construct per-level datasets $X^{(\ell)} = \{x_i^{(\ell)}\}_{i=1}^N$ for $\ell = 1,\dots,L$, ordered from the coarsest ($\ell = 1$) to the finest ($\ell = L$). Fix a reference level $\ell_0$ at which the data are given (so $x_i^{(\ell_0)} = x_i$). We generate neighboring levels using restriction and prolongation operators $R_{\ell\to\ell-1}$ and $P_{\ell\to\ell+1}$:

$$x_i^{(\ell-1)} = R_{\ell\to\ell-1}\, x_i^{(\ell)}, \qquad x_i^{(\ell+1)} = P_{\ell\to\ell+1}\, x_i^{(\ell)}.$$

Thus, for each $i$, the family $\{x_i^{(\ell)}\}_{\ell=1}^{L}$ represents the same underlying sample at different resolutions. In practice, $R_{\ell\to\ell-1}$ can be implemented by average pooling over local neighborhoods, while $P_{\ell\to\ell+1}$ can be implemented by interpolation.

We use a consistent encoder–decoder design across resolutions and train sequentially from coarse to fine. At level $\ell$, for a sample $x^{(\ell)} \in X^{(\ell)}$, with the encoder and decoder:

$$q_{\phi^{(\ell)}}(z \mid x^{(\ell)}), \qquad p_{\theta^{(\ell)}}(x^{(\ell)} \mid z),$$

and we optimize the level-$\ell$ ELBO

$$L_{\text{ELBO}}^{(\ell)} = \frac{1}{N}\sum_{i=1}^{N}\Big(\mathbb{E}_{q_{\phi^{(\ell)}}(z|x_i^{(\ell)})}\big[\log p_{\theta^{(\ell)}}(x_i^{(\ell)} \mid z)\big] - \mathrm{KL}\big(q_{\phi^{(\ell)}}(z \mid x_i^{(\ell)}) \,\|\, p(z)\big)\Big).$$

After training level $\ell$, we denote the weights by $(\tilde{\theta}^{(\ell)}, \tilde{\phi}^{(\ell)})$. To initialize level $\ell{+}1$, we transfer parameters via

$$(\theta_0^{(\ell+1)},\, \phi_0^{(\ell+1)}) \;:=\; T_{\ell\to\ell+1}\big(\tilde{\theta}^{(\ell)}, \tilde{\phi}^{(\ell)}\big),$$

where $T_{\ell\to\ell+1}$ denotes a parameter-transfer operator that maps the level-$\ell$ weights to an initialization at level $\ell{+}1$. We then optimize $L_{\text{ELBO}}^{(\ell+1)}$ from $(\theta_0^{(\ell+1)}, \phi_0^{(\ell+1)})$ to obtain $(\tilde{\theta}^{(\ell+1)}, \tilde{\phi}^{(\ell+1)})$. This process is summarized in Algorithm 5.1.

**Algorithm 5.1** Multigrid VAE Training

**Input:** Per-level datasets $\{X^{(\ell)}\}_{\ell=1}^{L}$, parameter-transfer operators $T_{\ell\to\ell+1}$
Initialize $(\theta_0^{(1)},\, \phi_0^{(1)})$
**for** $\ell = 1$ to $L$ **do**
  Train from $(\theta_0^{(\ell)},\, \phi_0^{(\ell)})$ on $X^{(\ell)}$ by maximizing $L_{\text{ELBO}}^{(\ell)}$ to obtain $(\tilde{\theta}^{(\ell)},\, \tilde{\phi}^{(\ell)})$
  **if** $\ell < L$ **then**
    Set $(\theta_0^{(\ell+1)},\, \phi_0^{(\ell+1)}) \leftarrow T_{\ell\to\ell+1}\big(\tilde{\theta}^{(\ell)},\, \tilde{\phi}^{(\ell)}\big)$
**Output:** final weights $(\tilde{\theta}^{(L)},\, \tilde{\phi}^{(L)})$

## 6. Numerical experiments.

We report numerical experiments to assess the effectiveness of the proposed multigrid training schemes. All experiments were run in Google Colab on a single NVIDIA Tesla T4 GPU. For multigrid GNNs, we assess efficiency and accuracy through convergence speed and final predictive performance relative to random initialization, examine stability through the effect of the guidance loss relative to vanilla multigrid, and test architectural generality by checking whether coarse-to-fine initialization consistently improves training across multiple GNN backbones. We evaluate these effects on two settings: BACE molecular graph classification and benchmark node classification datasets (Amazon Computers and OGBN-ArXiv).

We also compare multigrid initialization with a single-resolution baseline for a receptor-conditioned CVAE on CrossDocked2020. We evaluate generation quality using docking-based metrics (Vina energy and CNN affinity) under posterior and prior sampling.

### 6.1. BACE Dataset Experiments.

We first evaluate our approach on the BACE dataset, which contains experimental measurements of molecular activity against the human $\beta$-secretase 1 (BACE-1) protein. The dataset is sourced from MoleculeNet [54], a benchmark collection for molecular machine learning. The task is formulated as a graph classification problem, where each molecule is represented as a graph and labeled as either active ($y_i = 1$) or inactive ($y_i = 0$) based on its inhibitory effect on BACE-1.

Each molecule is initially represented as a SMILES string and then converted into a graph using RDKit and PyTorch Geometric. We model it as an undirected graph $G = (V, E)$, where nodes correspond to atoms, excluding implicit hydrogens, and edges represent chemical bonds. Node features capture detailed atomic properties such as atomic number, valence, hybridization state, and aromaticity. Edges are established between atoms that share an explicit chemical bond, with bond types such as single and double bonds encoded as edge features.

We adopt a two-layer Graph Isomorphism Network with Edge features (GINE) model as our GNN backbone. GINE builds upon the Graph Isomorphism Network (GIN), introduced by Xu et al. [55], which is one of the most expressive message passing architectures for graph level prediction. The GINE operator, introduced by Hu et al. [20], extends the GIN architecture by incorporating edge features into the aggregation process. This enhancement is particularly important in molecular graphs, where bond types carry significant semantic meaning. The update rule for each node $v$ at layer $l$ in GINE is given by

$$H_v^{(l)} = \mathrm{MLP}\left(\left(1 + \epsilon^{(l)}\right) H_v^{(l-1)} + \sum_{u \in \mathcal{N}(v)} \mathrm{ReLU}\left(H_u^{(l-1)} + e_{uv}\right)\right), \tag{6.1}$$

where $H_v^{(l)}$ is the embedding of node $v$ at layer $l$, $\mathcal{N}(v)$ denotes its neighbors, $e_{uv}$ is the edge feature between nodes $u$ and $v$, $\epsilon^{(l)}$ is a learnable or fixed scalar, and MLP is a multi-layer perceptron applied after aggregation.

We tested three training strategies using the GINE architecture. Method 1 (original training) trains on the original dataset for 1000 epochs from random initialization. Method 2 (subgraph pretraining) pretrains on upsampled subgraphs for 60 epochs, then fine tunes on the original dataset for 1000 epochs. Method 3 (layer-freezing) uses the same 60-epoch pretraining, then freezes the first GNN layer and trains on the original dataset for 200 epochs. Finally, all layers are unfrozen and training continues for 1000 epochs.

The results of these experiments are summarized in Figure 2. In the training curve, Method 3 achieves rapid convergence, with its BCE loss falling below 0.1 before epoch 400, whereas Method 1 only reaches the same threshold around epoch 800. In the validation curve, Method 3 consistently outperforms Method 1 and Method 2 across all epochs, maintaining the lowest BCE loss.

**6.2. Benchmark Node Classification Datasets.** We study two representative benchmark node-classification datasets. First, the Amazon Computers dataset from PyTorch Geometric is a co-purchasing network where nodes represent products and edges indicate items frequently bought together. Node features are derived from product metadata, and the classification task involves predicting product categories. It contains 13,752 nodes and 491,722 edges. We use 7,752 nodes for training, 3,000 for validation, and 3,000 for testing. To prevent data leakage during upsampling, test nodes are sampled from those in the full graph $G_K$ that do not appear in the finer subgraph $G_{K-1}$. We use a two-layer GCN with hidden dimension 16, dropout rate 0.5, and $L_2$ weight decay of $5 \times 10^{-4}$.

The second dataset, OGBN-ArXiv, is a large scale citation graph from the Open Graph Benchmark. Each node corresponds to an arXiv paper, with a 128-dimensional feature vector extracted from its title and abstract. Edges represent citation links, and the task is to classify papers into one of 40 subject areas. It contains 169,343 nodes and 1,166,243 directed edges. We use 10,000 nodes for validation, 10,000 for testing, and the remaining nodes for training. A two-layer GCN with hidden dimension 64 is used, and all other training hyperparameters are kept the same as in the Amazon Computers experiments.

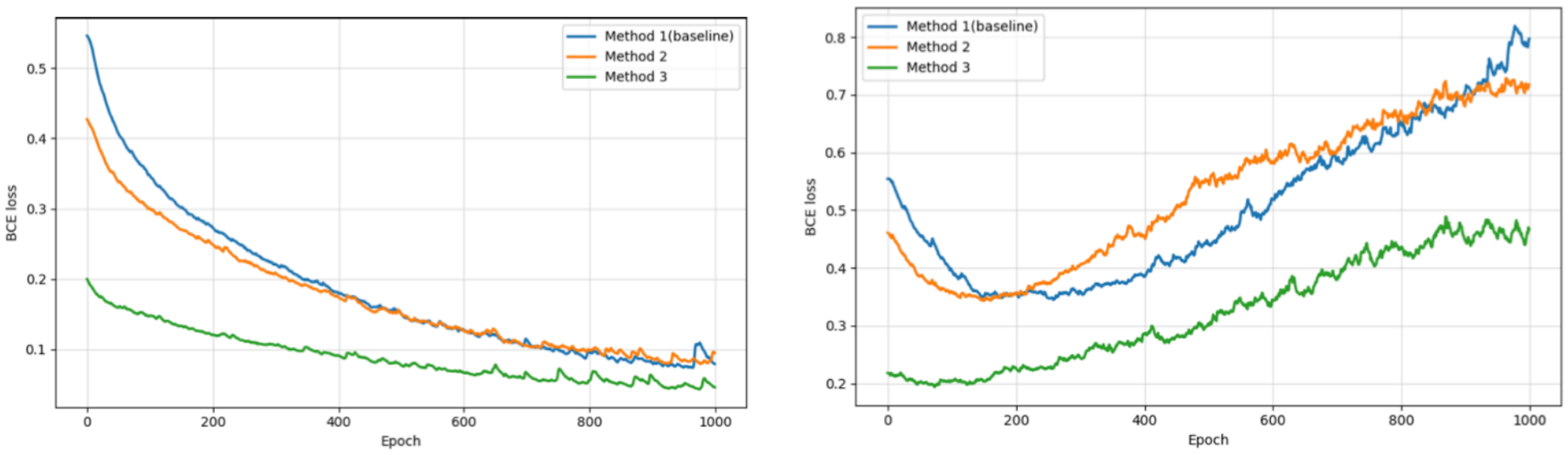

Fig. 2: Training BCE loss (left) and Validation BCE loss (right) curves for different training strategies on the BACE dataset.

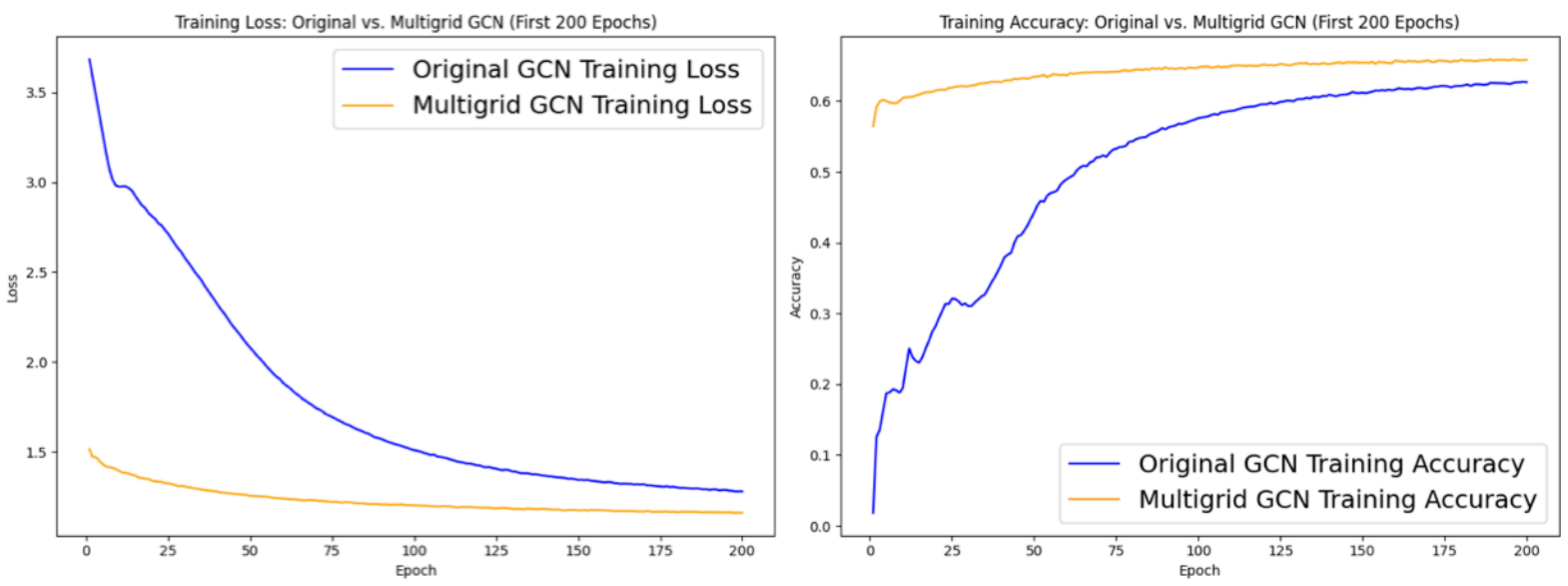

Fig. 3: Training loss and accuracy curves over the first 200 epochs on the OGBN-ArXiv dataset.

To quantify the effect of multigrid initialization, we repeat the training procedure over 10 random seeds, each producing a different multigrid hierarchy. This protocol assesses robustness across varying coarse-to-fine hierarchy constructions. For each run, we compare two GCN variants on the full graph: one trained from random initialization and the other initialized with parameters transferred from a GCN trained on a coarser resolution graph. Figures 3 and 4 illustrate the benefits of multigrid initialization. With multigrid initialization, training starts at a lower loss, converges faster, and maintains a lower validation loss over training.

To further assess training efficiency, we compare the average number of epochs to convergence under a 20-epoch early stopping criterion. That is, we stop training when the validation loss fails to decrease for 20 successive epochs. As shown in Table 1, multigrid initialization consistently leads to faster convergence while achieving slightly higher accuracy than random initialization. Moreover, using multiple starting nodes to construct a richer coarse subgraph yields even greater improvements, as shown in Table 2.

To evaluate the impact of our guidance loss, we perform a two-level multigrid procedure on the OGBN-ArXiv dataset. We first train on a coarse graph using an early-stopping window of 5 epochs, with a maximum of 200 epochs, followed by training on a finer graph with a 10-epoch patience window (also limited to 200 epochs). Across 10 different seeds, coarse-level training converges in an average of 52 epochs (0.46 seconds), while the finer-level model stops after 199 epochs (1.47 seconds). We then train on the full graph under various initialization strategies: random initialization (baseline), subgraph-based initialization without guidance ($\lambda = 0$), and subgraph-based initialization with guidance loss applied at different weights $\lambda$. Table 3 summarizes the results. Compared to random initialization, subgraph-based methods substantially reduce the number of epochs required for convergence while maintaining comparable or slightly improved accuracy. Adding guidance loss further accelerates convergence, with larger $\lambda$ values leading to shorter training durations at a modest cost to final accuracy. Figure 5 illustrates this trend through the upward trajectory of $-\ln(\text{epoch})$. These results confirm that guidance loss serves as an effective regularizer that enhances training stability and efficiency.

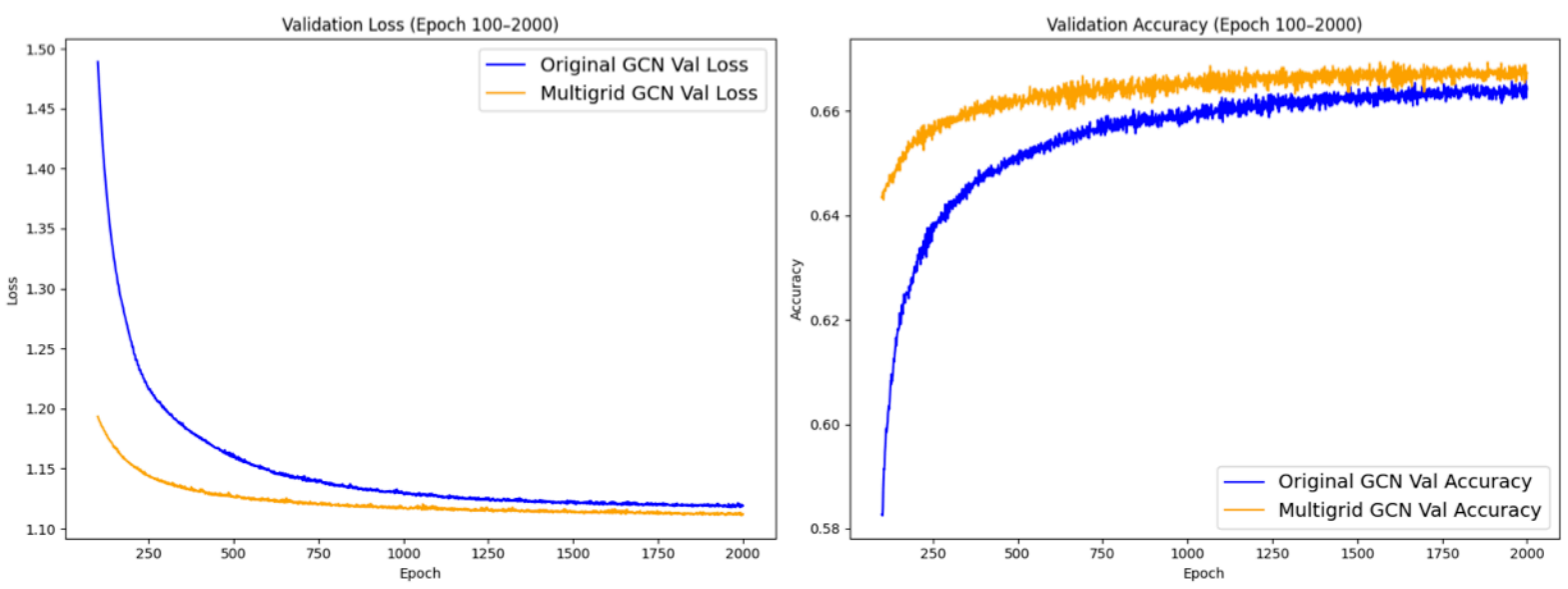

Fig. 4: Validation loss and accuracy curves over 2,000 epochs on the OGBN-ArXiv dataset.

Table 1: Accuracy and epoch comparison with early stopping for Amazon Computers dataset (1 starting node).

| Initialization | Accuracy [%] | Epochs |
|---|---|---|
| Multigrid init | $89.0 \pm 0.4$ | $331 \pm 24.3$ |
| Random init | $88.8 \pm 0.4$ | $497 \pm 39.9$ |

Table 2: Accuracy and epoch comparison with early stopping for Amazon Computers dataset (5 starting node).

| Initialization | Accuracy [%] | Epochs |
|---|---|---|
| Multigrid init | $88.7 \pm 0.4$ | $335 \pm 41.7$ |
| Random init | $88.0 \pm 0.7$ | $515 \pm 68.0$ |

**6.3. Receptor-Conditioned Ligand Generation.** Given a receptor binding pocket represented by the 3D coordinates and atom types of the pocket environment, our goal is to generate ligands that are geometrically compatible with the pocket and chemically valid. We represent the pocket and ligand with aligned atom-density grids and learn a conditional model $p_\theta(x \mid c)$, where $c$ denotes the receptor grid and $x$ the ligand grid.

We use the CrossDocked2020 dataset [10], a large collection of protein-ligand complexes with both re-docked and cross-docked poses (3D coordinates of receptors and ligands). In re-docking, each ligand is placed back into its original (cognate) receptor pocket, whereas in cross-docking, the same ligand is docked into other proteins with similar pockets, increasing diversity but adding noise. CrossDocked2020 contains 2,922 pockets, 18,450 pocket–ligand complexes, and 13,780 unique ligands, with over 22 million docked poses in total. Following common practice, we discard poses whose ligand RMSD exceeds 2 Å relative to the aligned reference pose and filter out molecules that fail RDKit sanitization.

Table 3: Performance on OGBN-ArXiv with different initialization schemes and guidance weights. All values are mean ± 95% CI over 10 seeds.

| Method | Epochs (Time [s]) | Accuracy [%] |
|---|---|---|
| Random init | $920 \pm 93.6\,(147.0 \pm 17.1)$ | $67.0 \pm 0.4$ |
| MG init $\lambda$=0 | $574 \pm 68.2\,(92.8 \pm 11.8)$ | $67.3 \pm 0.4$ |
| MG init $\lambda$=0.1 | $444 \pm 42.5\,(71.4 \pm 6.3)$ | $67.1 \pm 0.4$ |
| MG init $\lambda$=0.2 | $346 \pm 36.3\,(55.8 \pm 6.1)$ | $66.7 \pm 0.4$ |
| MG init $\lambda$=0.3 | $320 \pm 58.4\,(51.9 \pm 10.6)$ | $66.3 \pm 0.5$ |
| MG init $\lambda$=0.5 | $312 \pm 30.9\,(50.4 \pm 4.7)$ | $65.9 \pm 0.5$ |

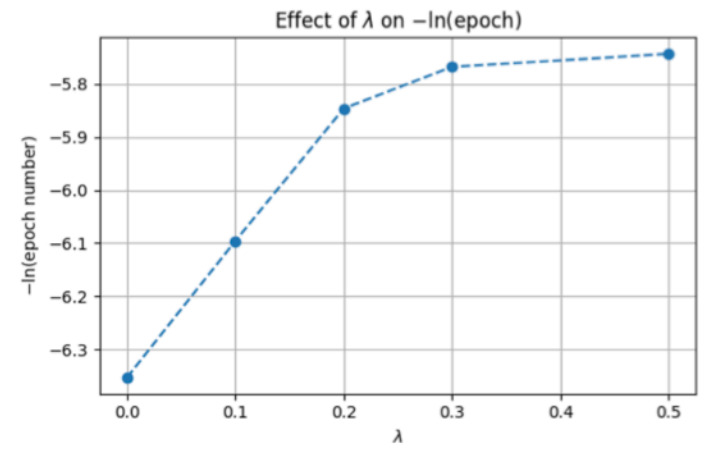

Fig. 5: $-\ln(\text{epoch})$ versus $\lambda$ on OGBN-ArXiv. Larger values indicate fewer epochs required to converge.

**6.3.1. Atom density grids.** To turn atoms into learnable features, we assign each atom a vector built from chemically meaningful properties: element, aromaticity, hydrogen bond donor status, hydrogen bond acceptor status, and formal charge. For a given atom $a$, we encode it with a one-hot vector for each property and then concatenate them:

$$t(a) = \big[\text{one-hot}_{\text{element}}(a) \parallel \text{one-hot}_{\text{aromatic}}(a) \parallel \text{one-hot}_{\text{donor}}(a) \\ \parallel \text{one-hot}_{\text{acceptor}}(a) \parallel \text{one-hot}_{\text{charge}}(a)\big].$$

We place a cubic grid around the binding site, typically centered at the ligand centroid. The box side length and voxel spacing determine spatial coverage and resolution. See Fig. 6 for an illustration. Following [38], we use a box of side 23.5 Å and a spacing of 0.5 Å, which yields a $48 \times 48 \times 48$ grid in our experiments.

Each atom is represented as a smooth Gaussian density centered at its coordinates. Let $d$ be Euclidean distance between a voxel center and an atom, and let $r$ be an atomic radius. The contribution of that atom to nearby voxels is given by

$$f(d,r) = \begin{cases} \exp\big(-2(d/r)^2\big), & d \le 1.5\,r, \\ 0, & d > 1.5\,r. \end{cases}$$

The radius is fixed at $r = 1$ Å for all atoms in this work. We start from a zero-valued grid and accumulate contributions from all atoms. For each atom $a$ in the box, we find all voxels within distance $1.5r$ of its position. In each voxel, we add $f(d,r)t(a)$ to the grid. Atoms outside the cubic box are ignored.

**6.3.2. Conditional Variational Autoencoder.** Given a receptor density grid, denoted as rec, we aim to sample ligand density grids, lig, from the conditional distribution $p(\text{lig} \mid \text{rec})$. We adopt a conditional Variational Autoencoder (CVAE), which introduces a latent variable $z$ of lower dimension to capture unobserved factors underlying receptor-ligand interaction patterns.

As shown in Fig. 7, the CVAE consists of two encoders and one decoder. The conditional encoder $g_\psi$ takes the receptor grid as input and produces a deterministic

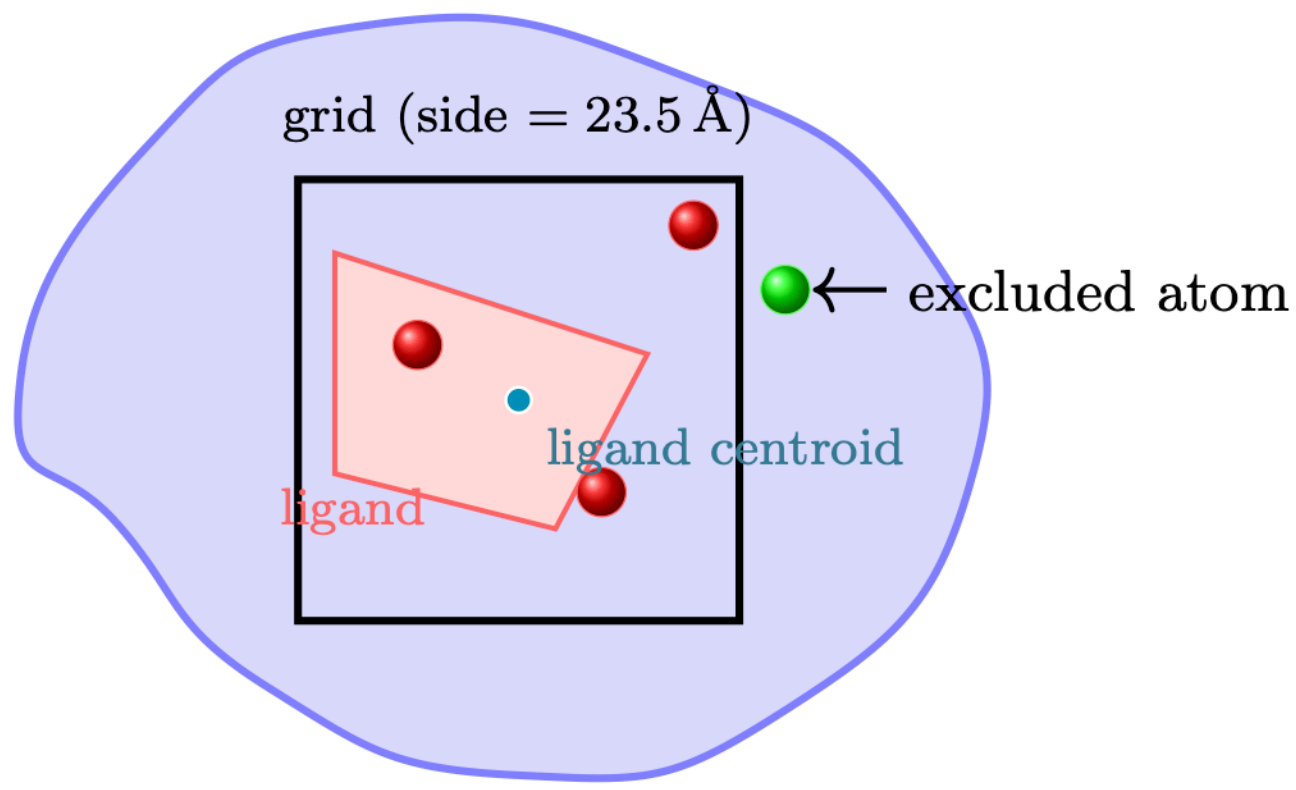

Fig. 6: 2D slice of the 3D density grid around a ligand. The square shows the cross-section of a cubic box centered at the ligand centroid (cyan). Red atoms contribute to voxel densities; atoms outside the box (green) are excluded.

conditioning vector

$$c = g_\psi(\text{rec}),$$

encoding the binding-site geometry and chemical context. The inference encoder $q_\phi$ takes $(\text{rec}, \text{lig})$ as input and outputs an approximate posterior over $z$:

$$q_\phi(z \mid \text{rec}, \text{lig}) = \mathcal{N}\big(\mu(\text{rec}, \text{lig}),\ \text{diag}(\sigma^2(\text{rec}, \text{lig}))\big).$$

Both encoders share the same 3D CNN architecture: four Conv3D blocks with channel widths $\{28, 56, 108, 216\}$, where the first three use $2 \times 2 \times 2$ average pooling, and the final block is followed by a fully connected layer of size 128. We then sample $z \sim q_\phi(z \mid \text{rec}, \text{lig})$ and generate a ligand grid conditioned on $c$:

$$\text{lig}_{\text{gen}} \sim p_\theta(\text{lig} \mid z, c).$$

The decoder $p_\theta$ mirrors the encoder design in reverse: it first maps the concatenated code $[z; c] \in \mathbb{R}^{256}$ through fully connected layers and reshapes it to a $6 \times 6 \times 6 \times 128$ latent grid, then applies four Conv3D blocks with channel widths $\{216, 108, 56, 28\}$, with $2 \times 2 \times 2$ nearest-neighbor upsampling after the first three. The final block is followed directly by a $3 \times 3 \times 3$ Conv3D layer that produces the ligand-density channels.

Training minimizes a standard CVAE objective with additional steric clash regularization. Let $\text{lig}_{\text{true}}$ denote the true ligand density grid and $\text{lig}_{\text{gen}}$ the decoder output. The reconstruction loss is

$$L_{\text{recon}} = \tfrac{1}{2} \left\| \text{lig}_{\text{true}} - \text{lig}_{\text{gen}} \right\|_2^2.$$

This $\ell_2$ loss corresponds to the negative log-likelihood of $p_\theta(\text{lig} \mid z, c)$ under an isotropic Gaussian observation model. Minimizing it encourages the decoded ligand density to match the ground truth ligand density at each voxel and channel.

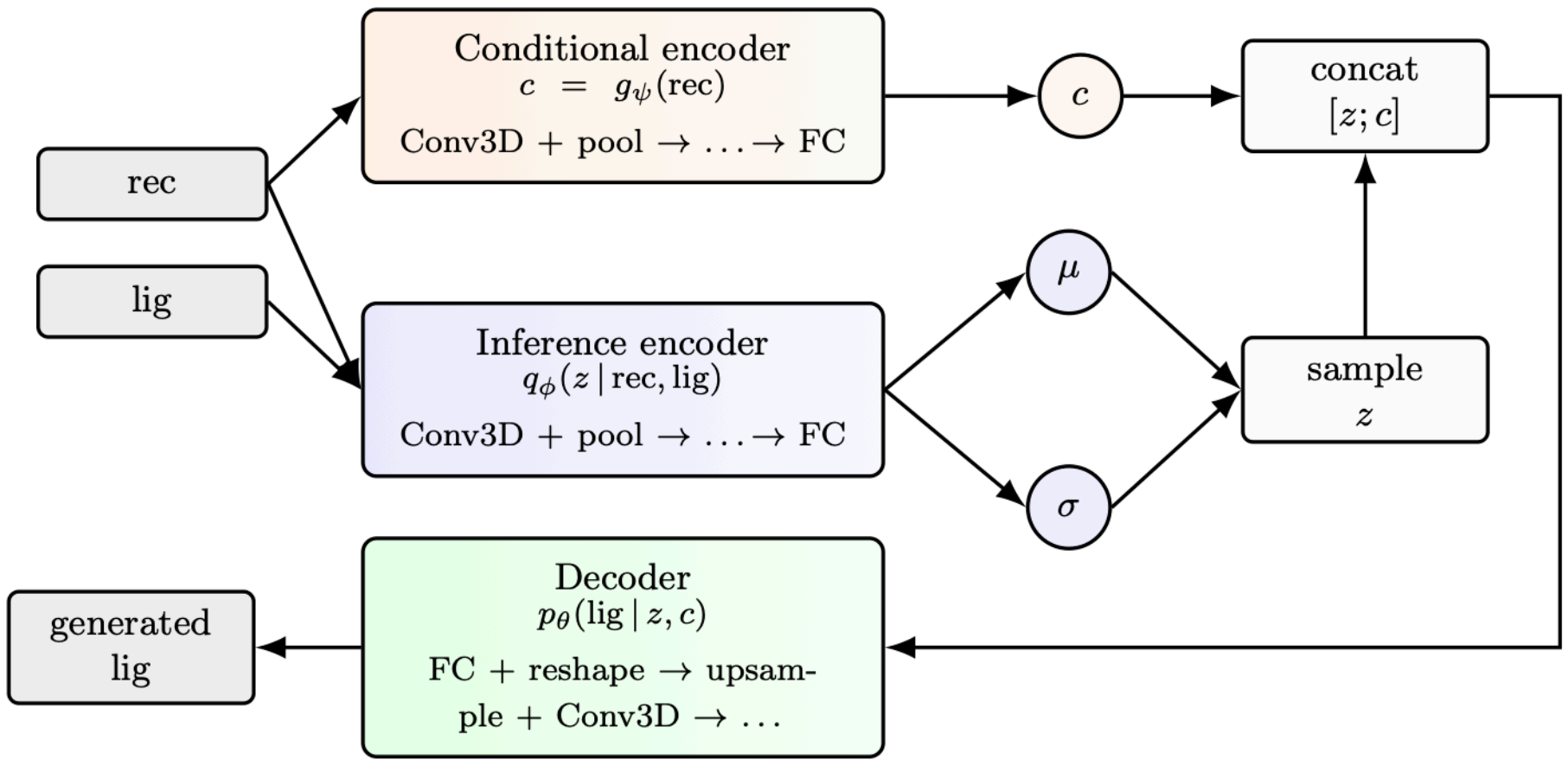

Fig. 7: Diagram of the receptor-conditioned CVAE. The inference encoder maps $(\text{rec}, \text{lig})$ to $(\mu, \sigma)$ for sampling $z$, while a conditional encoder produces a receptor code $c$ that conditions the decoder via concatenation.

The KL term encourages the approximate posterior $q_\phi(z \mid \text{rec}, \text{lig})$ to stay close to the standard normal prior $p(z) = \mathcal{N}(0, I)$:

$$L_{\text{KL}} = D_{\text{KL}}\big(q_\phi(z \mid \text{rec}, \text{lig}) \,\|\, p(z)\big).$$

To discourage steric clashes, we penalize overlap between the receptor density and the generated ligand density:

$$L_{\text{steric}} = \sum_{u,v,w} \Big(\sum_i \text{rec}_i(u, v, w)\Big)\Big(\sum_i \text{lig}_{\text{gen},i}(u, v, w)\Big),$$

where $i$ indexes the grid channels. The total loss is a weighted sum

$$L = \lambda_{\text{recon}} L_{\text{recon}} + \lambda_{\text{KL}} L_{\text{KL}} + \lambda_{\text{steric}} L_{\text{steric}}.$$

The decoder produces a multi-channel ligand density grid $\text{lig}_{\text{gen}}$, but downstream evaluation requires a discrete ligand with atom types, bonds and 3D coordinates. Converting a continuous density grid into a molecular structure is an ill-posed inverse problem with no closed-form solution. Following [38], we reconstruct molecules in two stages: (i) atom fitting to select atom types and 3D coordinates, and (ii) bond inference to assign bonds and hydrogens and enforce chemical validity.

*Atom fitting.* We formulate atom fitting as an optimization problem over atom types and 3D coordinates. Let $T$ denote the selected atom types and $C$ their 3D coordinates, and let $\mathcal{G}(T, C)$ be the procedure that converts $(T, C)$ into an atomic density grid. Atom fitting seeks

$$(T^*, C^*) = \arg\min_{(T,C)} \big\|\text{lig}_{\text{gen}} - \mathcal{G}(T, C)\big\|_2^2.$$

In practice, we use an iterative greedy refinement scheme. We propose candidate atom types and initial centers from voxels with high density in $\text{lig}_{\text{gen}}$. For each candidate, we locally refine its coordinates by gradient based optimization to reduce the $\ell_2$ reconstruction error $\|\text{lig}_{\text{gen}} - \mathcal{G}(T, C)\|_2^2$. At each iteration, we add one atom (type and position) corresponding to the candidate whose optimized placement yields the largest objective decrease, and we stop when no candidate provides a meaningful improvement.

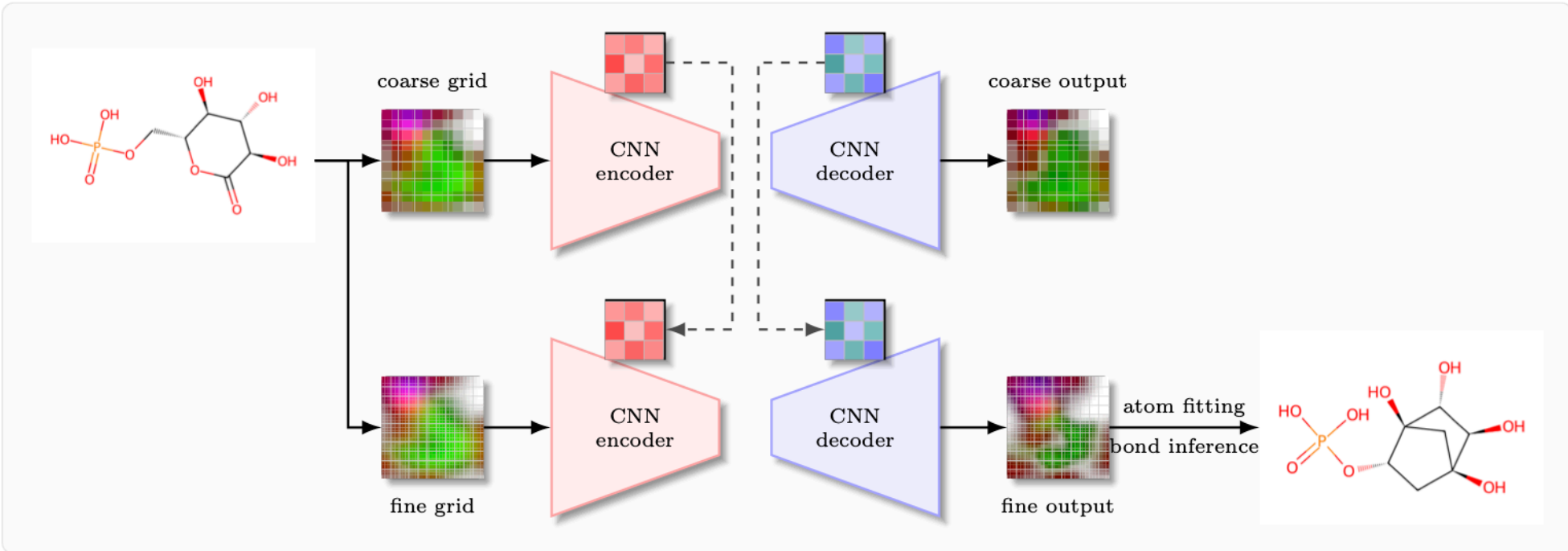

Fig. 8: Diagram of multigrid CVAE training. We first train a CVAE on a coarse voxel grid, then transfer shape compatible CNN weights to initialize the fine-resolution CVAE and continue training at the finer resolution. The resulting fine-grid density is converted to a discrete 3D molecule via atom fitting and bond inference.

*Bond inference.* Given the fitted atoms, we initialize a candidate bond graph by linking nearby atom pairs in 3D. We then filter the candidate bonds by enforcing valence constraints and excluding geometrically implausible connections. Finally, remaining open valences are satisfied by adding explicit hydrogens or adjusting bond orders where permitted, producing a discrete molecular graph with consistent 3D coordinates.

**6.3.3. Multigrid CVAE on atom density grids.** We now apply the general multigrid VAE scheme to the atom density grids. Figure 8 provides a schematic illustration of our approach. For each receptor–ligand complex, we fix the binding-site box (side length 23.5 Å) and choose a sequence of grid sizes $n_1 < n_2 < \cdots < n_L$. The voxel spacing at level $\ell$ is

$$h_\ell = \frac{23.5\,\text{Å}}{n_\ell - 1},$$

so each level represents the same physical box, differing only in resolution. At each level $\ell$, we apply the Gaussian density mapping to the receptor and ligand atoms, yielding multi-channel density tensors on the $n_\ell^3$ voxel lattice with a receptor grid $\text{rec}^{(\ell)} \in \mathbb{R}^{C_{\text{rec}} \times n_\ell \times n_\ell \times n_\ell}$ and a ligand grid $\text{lig}^{(\ell)} \in \mathbb{R}^{C_{\text{lig}} \times n_\ell \times n_\ell \times n_\ell}$. Figure 9 gives a toy 2D illustration of re-voxelizing the same atom on a coarse versus fine grid. In our experiments, we use a two-level hierarchy with $(n_1, n_2) = (16, 48)$, corresponding to spacings $h_1 = 23.5\,\mathring{A}/15 \approx 1.57\,\mathring{A}$ and $h_2 = 23.5\,\mathring{A}/47 = 0.50\,\mathring{A}$.

We use the same 3D CVAE design template at both resolutions, with a shared 128-dimensional latent space and identical block types across the encoder and decoder stacks. For multigrid training, we first train a coarse CVAE on $16^3$ atom-density grids, then use it to initialize a second CVAE on $48^3$ grids and continue training at the finer resolution. More specifically, our choice of $T_{\ell \to \ell+1}$ copies all convolutional weights in the receptor encoder, ligand encoder, and decoder whose parameter tensors have identical shapes in both models. Layers whose parameter shapes depend on the spatial resolution are not transferred and are initialized independently. In our architecture, this includes the encoder heads `FC 128` for $\mu$ and $\log \sigma$ and the decoder projection layer `FC 256`. Figure 10 illustrates this transfer scheme (decoder transfer is analogous and omitted for clarity).

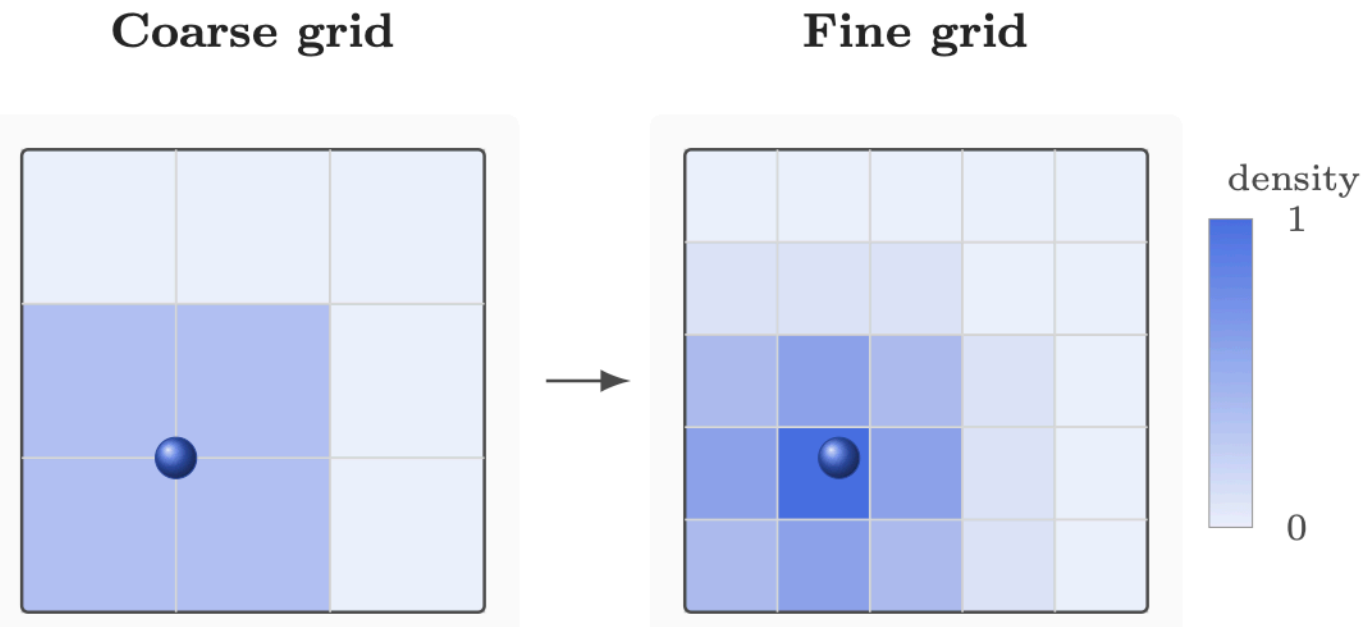

Fig. 9: Toy 2D slice of an atom-density channel voxelized over a fixed box at two resolutions. Colors indicate the Gaussian density value per voxel.

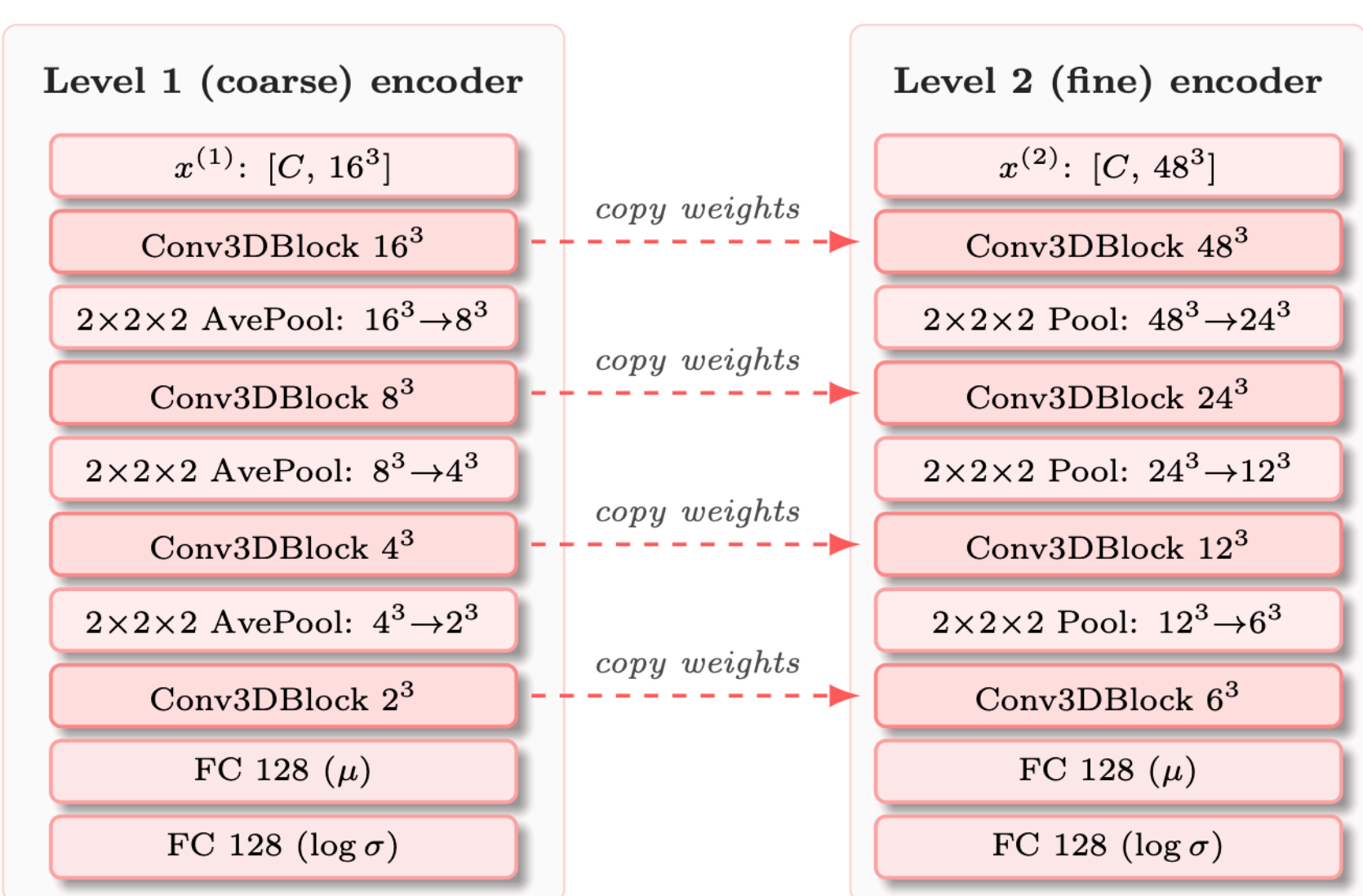

Fig. 10: Encoder weight transfer from a coarse CVAE trained on $16^3$ grids to a fine CVAE trained on $48^3$ grids. Each Conv3DBlock denotes a stack of four 3D convolutional layers. Dashed arrows indicate convolutional blocks whose weights are copied from the coarse model.

**6.3.4. Training and evaluation.** We use the same architecture for both a single-resolution $48^3$ CVAE trained from scratch and a multigrid $16^3 \to 48^3$ CVAE initialized by the weight-transfer scheme in Section 6.3.3. Both variants use a cyclical KL-weight schedule: in each cycle, we set $\lambda_{\mathrm{KL}} = 0.1$ for the first 45% of the cycle, increase it linearly from 0.1 to 1.6 over the next 20%, and hold $\lambda_{\mathrm{KL}} = 1.6$ for the remaining 35%. We repeat this schedule throughout training; the final cycle is truncated so that training ends at iteration 338,000. We optimize all models using RMSprop.

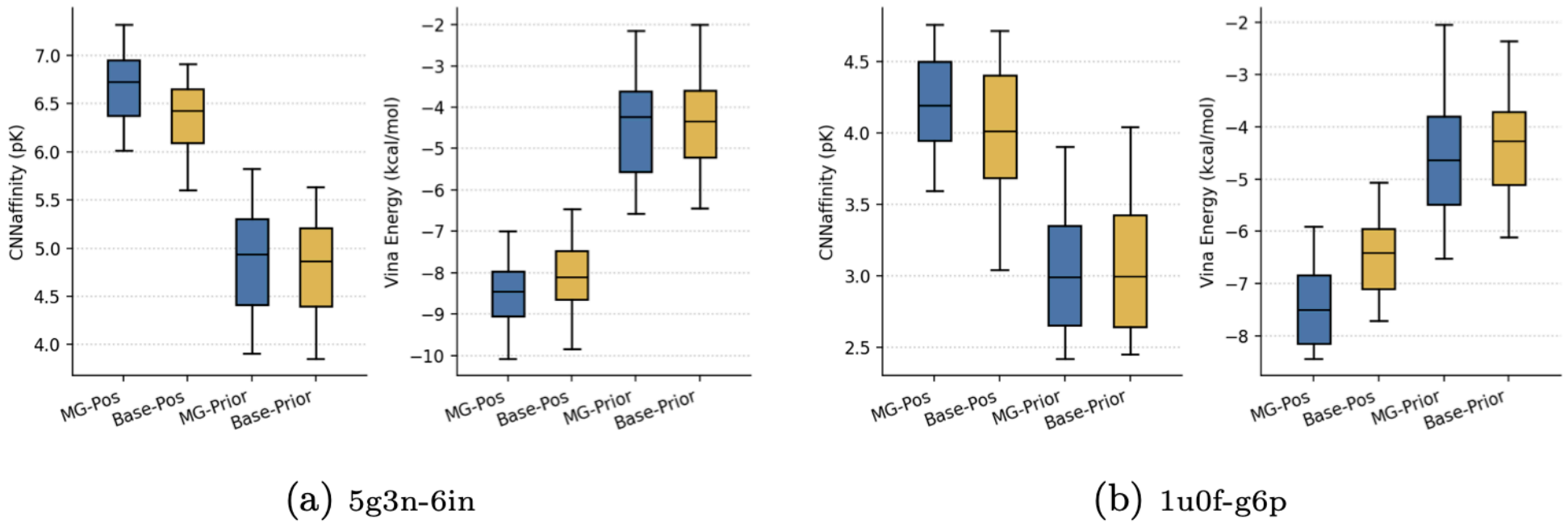

(a) 5g3n-6in (b) 1u0f-g6p

Fig. 11: Docking evaluation with GNINA on two CrossDocked2020 targets.

For the multigrid variant, we first train the coarse $16^3$ model for 1,000 iterations with learning rate $10^{-4}$ and batch size 32, since the coarse grids fit comfortably in memory. We then warm-start the $48^3$ model using the prolongation rule in Section 6.3.3 and continue training with learning rate $10^{-5}$ under the same KL schedule as the single-resolution baseline.

We generate ligands using three sampling modes.

- Posterior sampling: Given a receptor–ligand complex (rec, lig), we encode the pair to obtain posterior parameters $(\mu_{\text{post}}, \sigma_{\text{post}})$ and sample

$$z = \mu_{\text{post}} + \lambda_{\text{var}}\, \sigma_{\text{post}} \odot \varepsilon, \qquad \varepsilon \sim \mathcal{N}(0, I).$$

  Here $\lambda_{\text{var}}$ controls the sampling variability around the encoded ligand.
- Prior sampling: For prior sampling, we draw

$$z = \lambda_{\text{var}}\varepsilon, \qquad \varepsilon \sim \mathcal{N}(0, I),$$

  and decode using only the receptor condition.
- Mixed sampling: To interpolate between posterior- and prior-driven generation, we use a mixing coefficient $\alpha \in [0, 1]$ and sample

$$z = \alpha\, \mu_{\text{post}} + \lambda_{\text{var}}\big(\alpha\, \sigma_{\text{post}} + (1 - \alpha)\mathbf{1}\big) \odot \varepsilon, \qquad \varepsilon \sim \mathcal{N}(0, I).$$

  When $\alpha = 1$ this reduces to posterior sampling, and when $\alpha = 0$ it reduces to prior sampling. Intermediate $\alpha$ values control the degree of bias toward the reference ligand.

In all three modes, we compute a receptor embedding $c$ from the receptor encoder and decode the concatenated code $[z \parallel c]$ to obtain a ligand density grid. We then reconstruct a discrete 3D molecule using atom fitting followed by bond inference (Section 6.3.2).

To evaluate ligand quality, we use GNINA [32], a deep-learning–based molecular docking program, and report the Vina energy (kcal/mol; more negative is better) and the CNN affinity (pK; higher is better). For each generated ligand, we first relax the ligand geometry with UFF minimization in the binding site while keeping the receptor fixed, and then dock the relaxed ligand with GNINA to obtain both scores for the resulting poses. For each ligand, we keep the best-scoring pose and summarize score distributions over 100 samples across sampling modes and models. Figure 11 shows results on two CrossDocked2020 targets (5g3n-6in and 1u0f-g6p). Under posterior sampling, the multigrid initialized model produces higher CNN affinity and more favorable Vina energies than the single-resolution baseline, whereas under prior sampling the two models have largely overlapping distributions and both perform worse than in posterior sampling, since only posterior sampling is guided by the reference ligand through the encoded posterior.

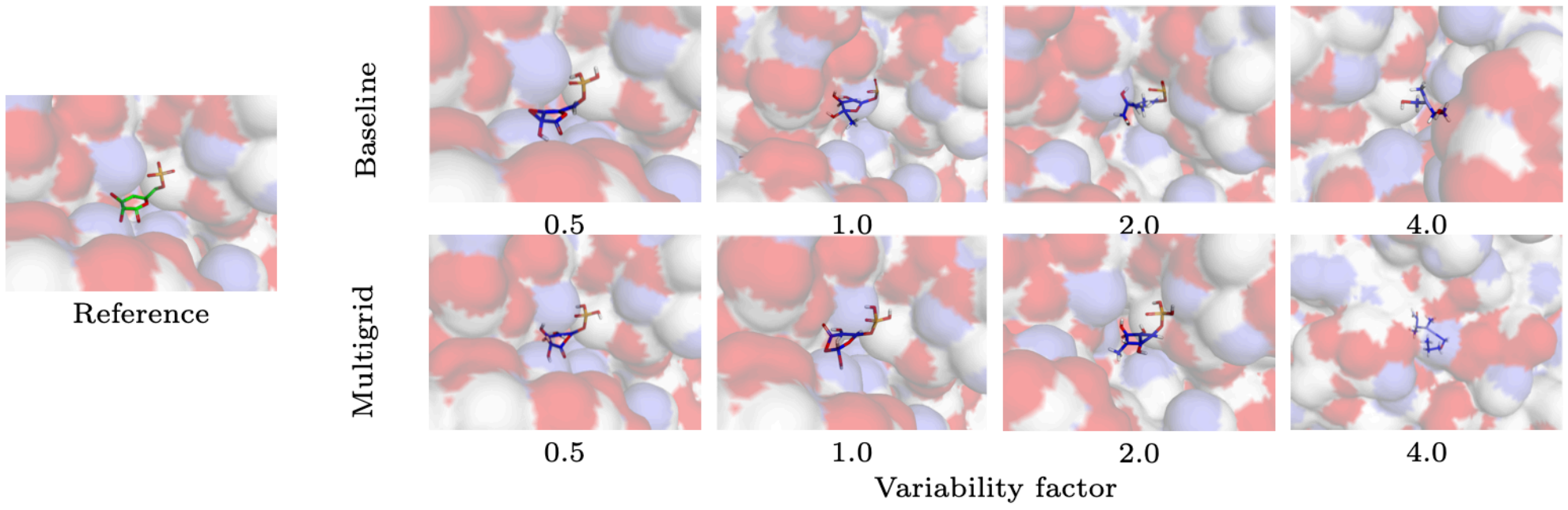

Fig. 12: Controlling the variability of generated molecules (1u0f-g6p). Left: reference ligand. Rows: baseline (top) and multigrid (bottom) posterior samples at matched variability factors.

We also study the effect of latent sampling on conditional generation by varying the posterior variability factor $\lambda_{\text{var}}$. Intuitively, $\lambda_{\text{var}}$ sets the sampling scale: smaller values keep samples near the posterior mean, while larger values allow larger deviations. Figure 12 illustrates the effect on a representative target: baseline (top row) and multigrid (bottom row), with matched $\lambda_{\text{var}}$. At low $\lambda_{\text{var}}$, generated ligands stay close to the reference pose with mostly local functional-group edits. Larger $\lambda_{\text{var}}$ yields progressively larger structural changes and greater pose variation in the pocket.

**7. Conclusion.** We presented a multigrid training framework that connects learning across resolutions in two settings. For graph neural networks, we introduced a coarse-to-fine training strategy that transfers parameters across graph resolutions and supports parameter sharing and layer freezing. Our analysis relates cross-level prediction, loss, and gradient discrepancies to the mismatch between fine and coarse operators, helping clarify when coarse warm starts are effective. For receptor-conditioned 3D ligand generation, we applied the same coarse-to-fine principle to conditional Variational Autoencoders on atom-density grids, transferring convolutional weights that are compatible across resolutions to initialize the fine model from the coarse one.

Across graph benchmarks, scientific regression tasks, and receptor-conditioned ligand generation, multigrid initialization improves training efficiency and achieves same level of accuracy compared with single-resolution training. Future work includes extending the theory to deeper architectures and stochastic optimization, as well as designing adaptive coarsening and prolongation operators that better preserve task relevant structure. On the generative side, it would be interesting to integrate multigrid initialization with diffusion models by transferring weights across spatial resolutions or across coarse-to-fine denoising schedules.

**Appendix A. Proof of Theorem 4.6.** Decomposing the fine loss into overlap and non-overlap parts, we have

$$L_f(w) = \frac{1}{2n}\left(\|r_f^U\|_2^2 + \|r_f^{U^c}\|_2^2\right) = \frac{\rho}{2m}\|r_f^U\|_2^2 + \frac{1}{2n}\|r_f^{U^c}\|_2^2.$$

Hence

$$|L_f(w) - L_c(w)| \le \frac{1}{2m}\left|\,\rho\|r_f^U\|_2^2 - \|r_c\|_2^2\,\right| \;+\; \frac{1}{2n}\|r_f^{U^c}\|_2^2.$$

Writing $\rho\|r_f^U\|^2 - \|r_c\|^2 = \rho(\|r_f^U\|^2 - \|r_c\|^2) - (1-\rho)\|r_c\|^2$, we have

$$\frac{1}{2m}\left|\,\rho\|r_f^U\|^2 - \|r_c\|^2\,\right| \le \frac{\rho}{2m}\left(\|r_f^U\| + \|r_c\|\right)\|r_f^U - r_c\| \;+\; \frac{1-\rho}{2m}\|r_c\|^2.$$

Observe that on $U$, the residuals differ by

$$r_f^U - r_c = Q^\top(\hat{Y}_f - Y) - (\hat{Y}_c - Y_c) = Q^\top\hat{Y}_f - \hat{Y}_c,$$

since $Y_c = Q^\top Y$. By Lemma 4.5,

$$\|r_f^U - r_c\|_2 = \|Q^\top\hat{Y}_f - \hat{Y}_c\|_2 \;\le\; L_\sigma\,\delta\,\|X\|_2\,\|w\|_2.$$

Plugging this into the previous bound yields the desired inequality.

**Appendix B. Proof of Lemma 4.7.** Since $D_{ii} = \sum_j \bar{A}_{ij} \ge 1$ for all $i$, the diagonal matrices $D^{-1}$ and $D^{-1/2}$ are well-defined. Set

$$B := D^{-1}\bar{A}.$$

Since $\bar{A}$ has nonnegative entries, $B$ also has nonnegative entries. Moreover, for each row $i$,

$$\sum_{j=1}^{n} B_{ij} \;=\; \sum_{j=1}^{n}\frac{\bar{A}_{ij}}{D_{ii}} \;=\; \frac{\sum_{j=1}^{n}\bar{A}_{ij}}{D_{ii}} \;=\; 1.$$

Hence the induced $\ell_\infty$ matrix norm satisfies

$$\|B\|_\infty = \max_i \sum_{j=1}^{n}|B_{ij}| = \max_i \sum_{j=1}^{n} B_{ij} = 1.$$

For any eigenpair $Bx = \lambda x$ with $x \ne 0$, we have

$$|\lambda|\,\|x\|_\infty = \|Bx\|_\infty \le \|B\|_\infty\|x\|_\infty = \|x\|_\infty,$$

which implies $|\lambda| \le 1$. Therefore every eigenvalue of $B$ lies in the closed unit disk.

Next, $B$ is similar to $\tilde{A}$:

$$\tilde{A} = D^{-1/2}\bar{A}D^{-1/2} = D^{1/2}(D^{-1}\bar{A})D^{-1/2} = D^{1/2}BD^{-1/2}.$$

Thus $B$ and $\tilde{A}$ have the same eigenvalues. Since $\tilde{A}$ is symmetric, all its eigenvalues are real, and hence they lie in $[-1,1]$. Finally,

$$\|\tilde{A}\|_2 = \max_i|\lambda_i(\tilde{A})| \le 1,$$

which completes the proof.

**Appendix C. Proof of Theorem 4.8.** By the chain rule, we have

$$\nabla L_f(w) = \frac{1}{n} X^\top A^\top D_f r_f, \qquad \nabla L_c(w) = \frac{1}{m} X^\top Q A_c^\top D_c r_c,$$

where $D_f := \mathrm{Diag}(\sigma'(AXw))$ and $D_c := \mathrm{Diag}(\sigma'(A_c X_c w))$. Decompose $r_f = Q r_f^U + r_f^{U^c}$ and define

$$\Delta_U := \frac{1}{n} X^\top A^\top D_f Q r_f^U - \frac{1}{m} X^\top Q A_c^\top D_c r_c, \qquad \Delta_{U^c} := \frac{1}{n} X^\top A^\top D_f r_f^{U^c}.$$

Then $\nabla L_f(w) - \nabla L_c(w) = \Delta_U + \Delta_{U^c}$.

Add and subtract the same coarse term with factor $1/n$ to separate overlap and averaging:

$$\Delta_U = \frac{1}{n} X^\top \Big( A^\top D_f Q r_f^U - Q A_c^\top D_c r_c \Big) \; + \; \Big( \tfrac{1}{n} - \tfrac{1}{m} \Big) X^\top Q A_c^\top D_c r_c.$$

Let $D_f^U := Q^\top D_f Q$. Since $QQ^\top$ and $D_f$ are diagonal, they commute; thus

$$Q D_f^U r_f^U = Q \, (Q^\top D_f Q) \, r_f^U = D_f \, (QQ^\top Q) r_f^U = D_f \, Q r_f^U.$$

On the overlap, using $D_f Q r_f^U = Q D_f^U r_f^U$ and since $\|A_c\|_2 \le 1$, $\|D_f^U\|_2, \|D_c\|_2 \le L_\sigma$, and $\sigma'$ is $L_{\sigma'}$-Lipschitz, we obtain

$$\begin{aligned}
&\big\| A^\top D_f Q r_f^U - Q A_c^\top D_c r_c \big\|_2 = \big\| (A^\top Q - Q A_c^\top) D_f^U r_f^U + Q A_c^\top (D_f^U r_f^U - D_c r_c) \big\|_2 \\
\le& \delta \, \|D_f^U\|_2 \, \|r_f^U\|_2 + \|D_f^U\|_2 \, \|r_f^U - r_c\|_2 + \|D_f^U - D_c\|_2 \, \|r_c\|_2 \\
\le& \delta \, L_\sigma \, \|r_f^U\|_2 + L_\sigma \, \|r_f^U - r_c\|_2 + \|D_f^U - D_c\|_2 \, \|r_c\|_2 \\
\le& \delta \, L_\sigma \, \|r_f^U\|_2 + \big( L_\sigma^2 + L_{\sigma'} \, \|r_c\|_2 \big) \, \delta \, \|X\|_2 \, \|w\|_2,
\end{aligned}$$

where the last line uses Lemma 4.5 and the $L_{\sigma'}$-Lipschitz property of $\sigma'$.

For the averaging piece, we have

$$\Big\| \Big( \tfrac{1}{n} - \tfrac{1}{m} \Big) X^\top Q A_c^\top D_c r_c \Big\|_2 \; \le \; \Big( \tfrac{1}{m} - \tfrac{1}{n} \Big) L_\sigma \, \|X\|_2 \, \|r_c\|_2.$$

Combining the overlap and the averaging term, we obtain

$$\begin{aligned}
\|\Delta_U\|_2 \le& \frac{\delta}{n} \, \|X\|_2 \Big[ L_\sigma \, \|r_f^U\|_2 + \big( L_\sigma^2 + L_{\sigma'} \, \|r_c\|_2 \big) \, \|X\|_2 \, \|w\|_2 \Big] \\
&+ \Big( \tfrac{1}{m} - \tfrac{1}{n} \Big) L_\sigma \, \|X\|_2 \, \|r_c\|_2.
\end{aligned}$$

For the fine-only part, note that $\|A\|_2 \le 1$ and $\|D_f\|_2 \le L_\sigma$. We have

$$\|\Delta_{U^c}\|_2 \; \le \; \frac{L_\sigma}{n} \, \|X\|_2 \, \|r_f^{U^c}\|_2.$$

Combining the estimates for $\Delta_U$ and $\Delta_{U^c}$ gives the stated bound.

**Appendix D. Proof of Theorem 4.9.** We prove the bound by a one-step recursion on $E_{1,j}$. Fix $j \in \{2, \dots, k\}$. Since $Q_{1:j} = Q_{j-1:j} \, Q_{1:j-1}$, we have $Q_{1:j}^\top = Q_{1:j-1}^\top Q_{j-1:j}^\top$. Add and subtract $Q_{1:j-1}^\top A_{j-1} Q_{j-1:j}^\top$ to get

$$\begin{aligned}
E_{1,j} &= Q_{1:j}^\top A_j - A_1 Q_{1:j}^\top \\
&= Q_{1:j-1}^\top Q_{j-1:j}^\top A_j - A_1 Q_{1:j-1}^\top Q_{j-1:j}^\top \\
&= Q_{1:j-1}^\top \big( Q_{j-1:j}^\top A_j - A_{j-1} Q_{j-1:j}^\top \big) + \big( Q_{1:j-1}^\top A_{j-1} - A_1 Q_{1:j-1}^\top \big) Q_{j-1:j}^\top \\
&= Q_{1:j-1}^\top E_{j-1,j} \; + \; E_{1,j-1} \, Q_{j-1:j}^\top.
\end{aligned}$$

Taking spectral norms and using $\|Q_{1:j-1}^\top\|_2 = \|Q_{j-1:j}^\top\|_2 = 1$ gives

$$\delta_{1,j} = \|E_{1,j}\|_2 \le \|E_{j-1,j}\|_2 + \|E_{1,j-1}\|_2 = \delta_{j-1,j} + \delta_{1,j-1}.$$

Iterating from $j = k$ down to $j = 2$ yields

$$\delta_{1,k} \le \delta_{k-1,k} + \delta_{1,k-1} \le \delta_{k-1,k} + \delta_{k-2,k-1} + \delta_{1,k-2} \le \cdots \le \sum_{\ell=1}^{k-1} \delta_{\ell,\ell+1},$$

which proves the claim.

**Appendix E. Proof of Theorem 4.10.** By Theorem 4.9,

$$\delta_{1,k} \le \sum_{\ell=1}^{k-1} \delta_{\ell,\ell+1} \le C \sum_{\ell=1}^{k-1} \tau_\ell^\alpha = C\tau_1^\alpha \sum_{\ell=0}^{k-2} q^{\alpha\ell} = C\,\tau_1^\alpha\, \frac{1-q^{\alpha(k-1)}}{1-q^\alpha}.$$

Since $T = \sum_{\ell=1}^{k-1} \tau_\ell = \tau_1 \frac{1-q^{k-1}}{1-q}$, we have $\tau_1 = T\frac{1-q}{1-q^{k-1}}$, which yields the equality in terms of $T$.

To show the bound tightens with $k$ when $T$ is fixed, set $r := q^{k-1} \in (0,1)$. The $T$-form of the bound can be written as

$$C\,T^\alpha \frac{(1-q)^\alpha}{1-q^\alpha}\, g(r), \qquad g(r) := \frac{1-r^\alpha}{(1-r)^\alpha}.$$

Consider $h(r) := \log g(r) = \log(1-r^\alpha) - \alpha \log(1-r)$. Differentiating gives

$$h'(r) = -\frac{\alpha r^{\alpha-1}}{1-r^\alpha} + \frac{\alpha}{1-r} = \alpha\Big(\frac{1}{1-r} - \frac{r^{\alpha-1}}{1-r^\alpha}\Big) = \alpha\frac{1-r^{\alpha-1}}{(1-r)(1-r^\alpha)} \;\ge\; 0,$$

since $r \in (0,1)$ and $\alpha > 1$ imply $r^{\alpha-1} \le 1$. Hence $h$ is increasing on $(0,1)$, and therefore $g(r)$ is also increasing on $(0,1)$. Because $r = q^{k-1}$ decreases as $k$ increases, $g(r)$ decreases with $k$, and the bound decreases in $k$.